\documentclass[11pt, a4paper, copyright, gdm]{google}

\usepackage[authoryear, sort&compress, round]{natbib}
\usepackage{nicefrac}
\usepackage{subcaption}
\usepackage{xspace}
\usepackage{makecell}      % multi-line table cells
\usepackage{placeins}      % \FloatBarrier
\usepackage{wrapfig}       % \begin{wraptable}/wrapfigure
\usepackage{multirow}
\usepackage[most]{tcolorbox}
\tcbuselibrary{breakable,skins}

\PassOptionsToPackage{table}{xcolor}  % no-op (xcolor already loaded);
\definecolor{spromptRed}{HTML}{B22222}
\definecolor{spromptRedBg}{HTML}{FDECEC}
\definecolor{spromptBlue}{HTML}{1F4E79}
\definecolor{spromptBlueBg}{HTML}{EAF2FA}
\definecolor{spromptTeal}{HTML}{0F6E5C}
\definecolor{spromptTealBg}{HTML}{E8F4F1}
\definecolor{spromptGray}{HTML}{4A4A4A}
\definecolor{spromptGrayBg}{HTML}{F4F4F4}

\newtcolorbox{spromptbox}[3][]{%
  enhanced,breakable,
  lines before break=0,
  colback=#3,colframe=#2,
  colbacktitle=#2,coltitle=white,
  fonttitle=\bfseries\small,
  title={#1},
  boxrule=0.6pt,arc=2pt,
  left=8pt,right=8pt,top=4pt,bottom=4pt,
  before skip=4pt,after skip=4pt,
  attach boxed title to top left={xshift=6pt,yshift=-3pt},
  boxed title style={boxrule=0pt,arc=2pt},
}

\usepackage{silence}
\newcommand{\msd}[2]{#1\,{\tiny$\pm$\,#2}}

\newcommand{\ourdata}{3DCodeBench\xspace}

\uselogo{}

\title{\ourdata{}: Benchmarking Agentic Procedural 3D Modeling Via Code}

\reportnumber{}

\author[1,3]{Yipeng Gao}
\author[1]{Lei Shu}
\author[1]{Genzhi Ye}
\author[1]{Xi Xiong}
\author[2]{Ameesh Makadia}
\author[1]{Meiqi Guo}
\author[3]{Laurent Itti}
\author[1]{Jindong~Chen}

\affil[1]{Google DeepMind}
\affil[2]{Google Research}
\affil[3]{University of Southern California}

\correspondingauthor{yipengga@usc.edu, leishu@google.com}

\paperurl{}

\begin{abstract}
Procedural 3D modeling through code is emerging as a versatile paradigm, offering deterministic, engine-ready, and precisely editable assets that neural 3D generators inherently lack.
Authoring such procedural content, however, demands deep expertise in 3D software APIs, parametric design, and code-level geometric reasoning.
In this paper, we propose \ourdata{}, a systematic benchmark for evaluating vision-language model (VLM) agents for procedural 3D generation in 3D modeling software. 
Specifically, \ourdata{} evaluates how effectively 12 advanced VLMs can serve as procedural 3D modelers by translating text and image references into procedural code for 3D modeling software. 
Recognizing that automated metrics may not fully capture the perceptual quality of 3D shapes, we build 3DCodeArena, a ranking platform based on pairwise human preferences over generated 3D outputs. 
From extensive evaluations and results, we observe that: 
(1) Failures mostly arise from API mismatches, while successful renders still suffer from disconnected or floating 3D geometric components.
(2) Test-time scaling, such as higher thinking budgets and multi-turn refinement, improves performance overall. 
Our findings highlight a critical need for high-quality procedural coding data to advance commercial VLMs. 
Furthermore, effective procedural 3D modeling requires a robust execution environment that provides high-fidelity feedback for iterative refinement. 
We release \ourdata{}, including the curated large-scale dataset of multimodal (text/image) prompts, procedural code, 3D object triplets, evaluation protocol, and the public 3DCodeArena platform as a foundational toolkit for exploring VLM-based procedural 3D modelers.

\vspace{3mm}
Project Page: \href{https://3dcodebench.com}{3dcodebench.com}
\end{abstract}

\begin{document}

\maketitle

% ----- teaser ---------------------------------------------------------
\begin{figure}[t!]
  \centering
  \includegraphics[width=0.99\linewidth]{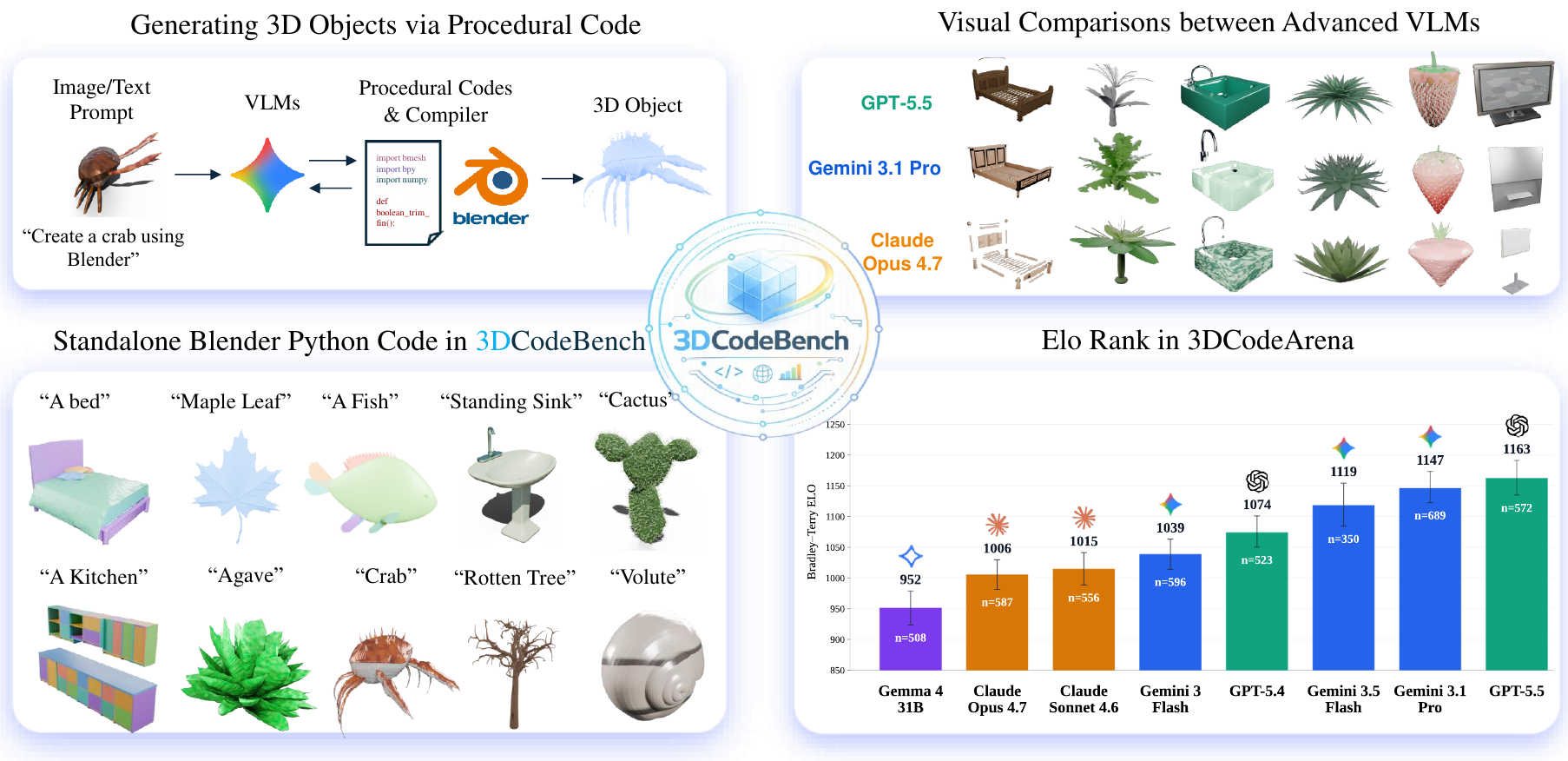}
  \caption{\textbf{Overview of \ourdata{}.} \textbf{(Top Left)} We explore the potential of Vision-Language Models to generate 3D objects via Procedural Code. \textbf{(Bottom Left)} The benchmark offers diverse, seed-addressable procedural test cases across a range of semantic categories. \textbf{(Top Right)} Qualitative comparisons of frontier VLMs (GPT-5.5, Gemini 3.1 Pro, Claude Opus 4.7) reveal varying capabilities in geometric reasoning. \textbf{(Bottom Right)} Alongside automated metrics, \emph{3DCodeArena} establishes Elo rankings via pairwise human-preference evaluations.}
  \label{fig:teaser}
\end{figure}

\section{Introduction}
\label{sec:intro}

\vspace{-0.8em}
\begin{quote}
\itshape\small
``Is God a Programmer, Not a Mathematician?''\hfill\normalfont\textmd{--- Gregory J. Chaitin}
\end{quote}
\vspace{-0.8em}

Procedural 3D modeling via coding is a critical pillar of modern digital creation~\citep{sidefx_houdini,idv_speedtree,adobe_substance3d_designer,blender_geometrynodes,esri_cityengine}, driving immense commercial value across gaming, industrial design, and high-fidelity simulation environments for robotics training~\citep{denninger2019blenderproc,greff2022kubric,deitke2022procthor,raistrick2023infinigen,raistrick2024infinigenindoors}.
Nevertheless, authoring these 3D procedural assets remains a labor-intensive and formidable challenge.
It demands that human designers have deep expertise in domain-specific coding syntax to manually define intricate meshes, complex geometric shapes, and realistic textures, and to perform precise parametric tuning to ensure physical plausibility.
To address these bottlenecks, the rapid advancement of Vision-Language Models (VLMs) and coding agents has paved the way for automating this complex development process.

Recognizing this automation potential, the community has been actively exploring ways to enable VLM agents in 3D creative artistry.
For instance, Anthropic announced the \emph{Claude for Creative Work}~\citep{anthropic2026creative} initiative to drive 3D modeling software like Blender via Python APIs.
Prior to this, an active ecosystem of Model Context Protocol (MCP) servers and function-calling agents~\citep{blender_mcp_official,ahujasid_blendermcp,elithril_blender_kiln,ra100_blender_claude_plugin,saofund_llm_blender_agent,minihellboy_claude_blender} has already turned frontier VLMs into natural-language drivers for end-to-end procedural modeling.
On the research side, a line of works~\citep{sun20243dgpt,hu2024scenecraft,lu2025ll3m,yin2026viga,yang2024holodeck,ling2025scenethesis} also builds LLM-driven agents to author procedural code or compose retrieved assets.
However, the community lacks a standardized, reliable benchmark for examining the model capabilities.

Evaluating the proficiency of VLMs as procedural modelers remains an open challenge due to four core limitations in existing methodologies. 
First, there is a severe lack of aligned procedural data. While repositories like ShapeNet~\citep{chang2015shapenet} and Objaverse~\citep{deitke2023objaversexl} provide vast collections of static meshes, they lack the underlying procedural code needed to measure generative agency. 
Second, current environments fail to capture real-world geometric complexity. 
For instance, BlenderGym~\citep{gu2025blendergym} focuses heavily on editing pre-built scenes rather than on from-scratch generation, and VoxelCodeBench~\citep{zheng2026voxelcodebench} limits construction to simplistic voxel grids. 
Third, existing frameworks ignore the iterative nature of 3D design by relying on single-shot evaluations and omitting the agentic refinement loops that are crucial to real workflows. 
Finally, the field lacks standardized evaluation metrics. 
Current models are tested on ad hoc prompts without a shared reference set, and the community lacks comprehensive perception metrics alongside a systematic 3D code arena for human preference voting.
Consequently, the research community lacks a comprehensive, standardized benchmark to rigorously measure the performance of frontier VLMs at generating 3D assets from code.

In this paper, we introduce \ourdata{} (Figure~\ref{fig:teaser}), a comprehensive benchmark and evaluation framework that directly addresses these four limitations. 
First, to address data scarcity, we extract procedural factories from Infinigen~\cite{raistrick2023infinigen,raistrick2024infinigenindoors} and process them using an agentic curation pipeline with human verification.
This yields a high-quality dataset of 26K text/image prompts $\leftrightarrow$ standalone code $\leftrightarrow$ 3D object triplets across 212 categories (Figure~\ref{fig:teaser}, bottom left).
Second, our dataset captures real-world geometric complexity rather than constructing objects using simple primitive shapes. 
Classes such as flying bird, crab, and dragonfly---each averaging over 400 lines of code---push models far beyond simplistic shape primitives (Figure~\ref{fig:teaser}, top right).
Third, to reflect the iterative nature of 3D design, our framework moves beyond single-shot prompting by treating VLMs as active \emph{agents}. 
We systematically evaluate their abilities in controlled multi-turn scenarios that include execution-error feedback, retries, visual self-critique, and API documentation augmentation.
Finally, we establish a standardized and robust evaluation protocol. We extensively analyze 12 advanced VLMs (Gemini~\citep{gemini31pro_blog}, Claude~\citep{anthropic2025claude4}, and GPT~\citep{openai2025gpt5} series) using comprehensive automated metrics spanning executability, perceptual view similarity (SigLIP-2~\citep{tschannen2025siglip2}, DINOv3~\citep{simeoni2025dinov3}), and 3D geometric alignment (Chamfer distance~\citep{fan2017psgn}, Uni3D~\citep{zhou2024uni3d}). To complement these metrics, we launch \emph{3DCodeArena} (Figure~\ref{fig:teaser}, bottom right), a public arena that collects pairwise human preferences to produce reliable Elo rankings.

In summary, \ourdata{} provides a comprehensive benchmark for evaluating the procedural 3D generation capabilities of VLM agents.
Through extensive evaluations, we mainly find: \emph{(1) Physical Plausibility remains the primary bottleneck beyond Executability:} Models frequently produce disconnected parts and incorrect structural alignments, revealing a critical lack of physical-world understanding. 
\emph{(2) Test-Time Scaling:} Multi-turn refinement improves performance using deterministic execution feedback from Blender.
This demonstrates that designing an appropriate agentic harness to process environmental feedback is crucial to unlocking the model's full potential.
Our contributions are summarized as:
\begin{itemize}[leftmargin=1.5em,itemsep=2pt,topsep=2pt]
  \item \textbf{\ourdata{}:} A standardized, VLM-driven procedural 3D modeling benchmark. Spanning a diverse range of 212 object classes and 26K 3D object--code pairs, it is curated via the proposed agentic pipeline with human feedback and paired with comprehensive evaluation protocols. 
  \item \textbf{3DCodeArena:} A systematic human-preference platform designed to evaluate the perceptual quality and aesthetic appeal of generated 3D shapes through pairwise Elo rankings.
  \item \textbf{Extensive VLM Analysis \& Insights:} A holistic evaluation of 12 advanced VLMs that identifies critical failure modes and establishes multi-turn agentic refinement as the primary driver for successful procedural 3D generation. 
\end{itemize}

\section{Related Work}
\label{sec:related}

\textbf{Procedural and Agentic 3D Generation.}
Automated 3D content creation has progressed from static repositories (e.g., ShapeNet~\citep{chang2015shapenet}, ABO~\citep{collins2022abo}, Thingi10K~\citep{zhou2016thingi10k}, Objaverse~\citep{deitke2023objaverse,deitke2023objaversexl}, OmniObject3D~\citep{wu2023omniobject3d}, ScanNet~\citep{dai2017scannet}, Cap3D~\citep{luo2023cap3d}) to dynamic procedural pipelines such as Infinigen~\citep{raistrick2023infinigen,raistrick2024infinigenindoors,joshi2025infinigensim}, BlenderProc~\citep{denninger2019blenderproc}, Kubric~\citep{greff2022kubric}, and ProcTHOR~\citep{deitke2022procthor}, which utilize hand-written code or templates.
Early methods learned shape specifications~\citep{sharma2018csgnet,jones2020shapeassembly,jones2023shapecoder}. In contrast, recent systems such as 3D-GPT~\citep{sun20243dgpt}, SceneCraft~\citep{hu2024scenecraft}, and LL3M~\citep{lu2025ll3m} leverage vision-language models (VLMs) to directly generate production-grade procedural scripts, while other works focus on articulated object generation~\citep{zhou2026articraft,joshi2025infinigensim, le2025articulate-anything}.
To reduce the complexity of raw application programming interfaces (APIs), Proc3D~\citep{raji2026proc3d} generates simplified procedural graphs. Simultaneously, an iterative agentic paradigm is emerging. For example, 3D-Generalist~\citep{sun20253dgeneralist} frames 3D synthesis as a sequential decision-making process based on visual observations, and CADCodeVerify~\citep{alrashedy2024cadcodeverify} demonstrates VLM self-correction of CAD code through visual inspection. At the scene level, Holodeck~\citep{yang2024holodeck} and Scenethesis~\citep{ling2025scenethesis} compose layouts for retrieved assets, while VIGA~\citep{yin2026viga} utilizes a write-run-render loop for reconstruction.

\textbf{Evaluating Procedural 3D Generation.}
General-purpose code benchmarks such as HumanEval~\citep{chen2021humaneval}, MBPP~\citep{austin2021mbpp}, and CodeContests~\citep{li2022alphacode} rely on unit tests.
Existing benchmarks target specific aspects of procedural 3D generation capability. 
BlenderGym~\citep{gu2025blendergym} focuses on scene editing rather than generation from scratch. VoxelCodeBench~\citep{zheng2026voxelcodebench} evaluates single-shot inference on low-complexity voxel structures. SceneScript~\citep{avetisyan2024scenescript} emphasizes layout prediction instead of executable asset generation. 
3DGen-Bench~\citep{zhang20253dgenbench} collects human preferences for neural text-to-3D outputs but does not include a code modality. 
CADBench~\citep{du2024blenderllm} evaluates text-conditioned object quality but is limited to low-complexity objects.
In contrast, \ourdata{} conceptualizes procedural 3D evaluation as an agentic task.
\ourdata{} assesses vision-language models both as single-shot generators and as iterative agents that author, execute, and visually refine complex Blender scripts.
\section{3DCodeBench: Towards 3D Generation via Writing Procedural Code}
\label{sec:pipeline}

\subsection{Task Definition}
\label{sec:task}
Procedural 3D generation requires a policy for synthesizing executable code that a 3D software runtime compiles into a target object.
Formally, given a condition $c$ (text and optional reference images), a policy $\pi$ produces a script $f_\pi = \pi(c)$ that a deterministic operator $\mathcal{E}$ executes into a mesh $M_\pi = \mathcal{E}(f_\pi)$; we instantiate $(\pi, \mathcal{E})$ on Blender~5.0 as a representative platform, making $f_\pi$ a Blender Python script, but the formulation is software-agnostic.
Each task is a triplet $(c, f^{\star}, M^{\star})$ of prompt, reference script, and ground-truth mesh, scored by a binary \emph{executability} indicator $\mathbb{I}[\mathcal{E}(f_\pi) \neq \emptyset]$ and continuous \emph{mesh-grounded similarity} $\mathcal{D}(M_\pi, M^{\star})$.
To probe agentic capabilities we permit $T \ge 1$ refinement iterations: at step $t$ the policy updates $f_\pi^{(t)}$ from execution logs or visual feedback, and the final evaluation uses $M_\pi^{(T)}\!=\!\mathcal{E}(f_\pi^{(T)})$, covering both single-shot ($T{=}1$) and multi-turn ($T{>}1$) settings.

\subsection{Agentic Data Curation Pipeline with Human Feedback}
\label{sec:pipeline1}

\begin{figure}[tbp]
  \centering
  \includegraphics[width=0.97\linewidth]{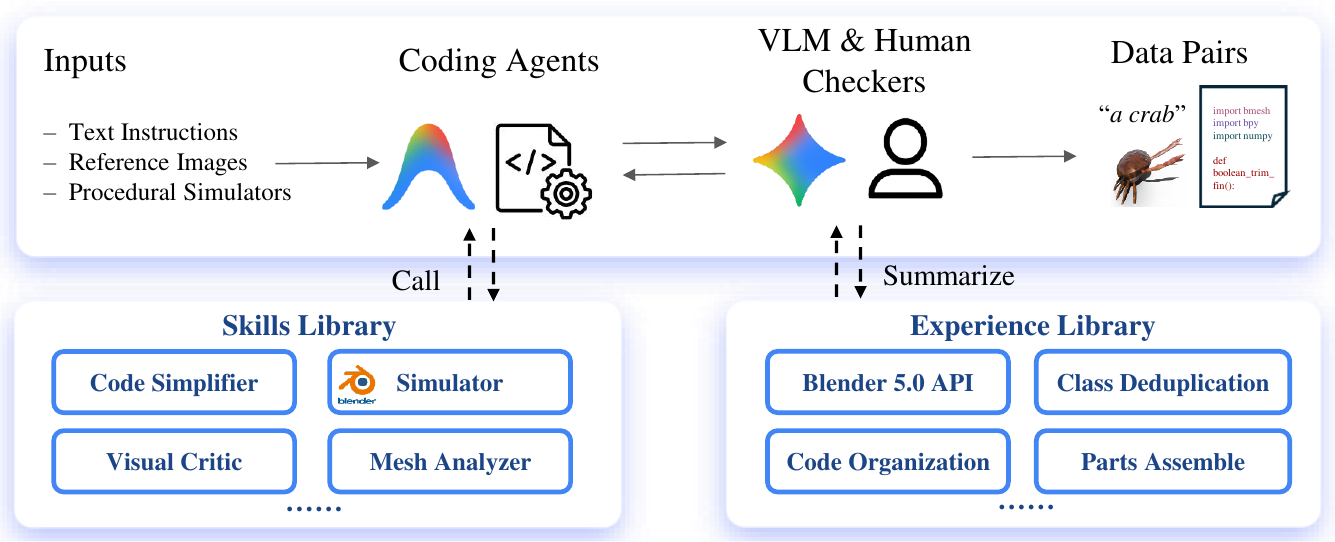}
  \caption{\textbf{Agentic data-curation pipeline.} A VLM-driven agent leverages a structured knowledge base to transform complex procedural factories into standalone Blender Python scripts via API migration, geometric validation, and code simplification. Visual fidelity is maintained through an iterative refinement loop using multi-view renders. Finally, every generated \emph{(prompt, code, mesh)} triplet undergoes strict \textbf{human-in-the-loop} verification to guarantee the highest benchmark quality.}
  \label{fig:pipeline}
\end{figure}

To construct \ourdata{} at scale, we introduce an automated curation pipeline (Figure~\ref{fig:pipeline}) that transforms deeply nested procedural factories from Infinigen into standalone Python scripts. The pipeline processes text instructions, reference images, and raw source code, facilitating a continuous feedback loop between coding agents and verification tools. To address the complexity of this software engineering task and reduce manual intervention, agents use two core components: a \textbf{Skills Library} for execution feedback and an \textbf{Experience Library} for retrieving established solutions.

\textbf{Skills Library.} Agents iteratively refine scripts by invoking specialized tools that provide objective feedback. The \textbf{Code Simplifier} automatically reduces long, deeply nested source code into clean, standalone scripts while strictly preserving the original 3D shape. The \textbf{Simulator} executes the generated code in a sandboxed Blender~5.0 environment to catch runtime errors and extract mesh data. To assess appearance, the \textbf{Visual Critic} (a VLM) compares multi-view renders of the generated object against the original reference, guiding the agent to correct visual discrepancies. Additionally, the \textbf{Mesh Analyzer} checks for structural issues---such as invalid geometry, non-manifold artifacts, or abnormally high vertex counts---to ensure the resulting 3D model is well-formed and physically plausible.

\textbf{Experience Library.} To avoid repeating the same mistakes, the pipeline builds a shared, continuously expanding knowledge base. When VLMs or human checkers identify recurring issues, they document successful strategies in this library, which consists of four core modules. \textbf{Class Deduplication} maintains a dynamic list of processed categories to filter out redundant classes, ensuring the high diversity and quality of the curated data. \textbf{Parts Assembly} provides structured templates that guide the modeling and integration of individual geometric parts into a coherent, holistic object. The \textbf{Blender 5.0 API} module continuously catalogs syntax changes and migration rules from older Blender versions, enabling agents to resolve deprecation errors preemptively. Finally, \textbf{Code Organization} enforces standardized stylistic conventions and architectural layouts for the generated Python scripts. 

\textbf{Human-in-the-Loop Verification.} Although the agentic loop automates most of the curation process, human oversight serves as the final quality control and fallback mechanism. Annotators manually review generated samples to verify reliable execution, the semantic accuracy of captions (using Gemini~3.1 Pro~\citep{gemini31pro_blog}), and visual alignment with reference images. If coding agents consistently fail to produce satisfactory results, a human expert intervenes by providing targeted textual feedback on specific object parts or by visually annotating rendered images to guide the agents. Only data pairs that pass this rigorous audit are included in the benchmark, ensuring high-fidelity \emph{(prompt, code, mesh)} triplets.

\subsection{Statistics of \ourdata{} and Curated 3D Code Data}
\label{sec:stats}

\begin{figure}[tbp]
  \centering
  \includegraphics[width=0.99\linewidth]{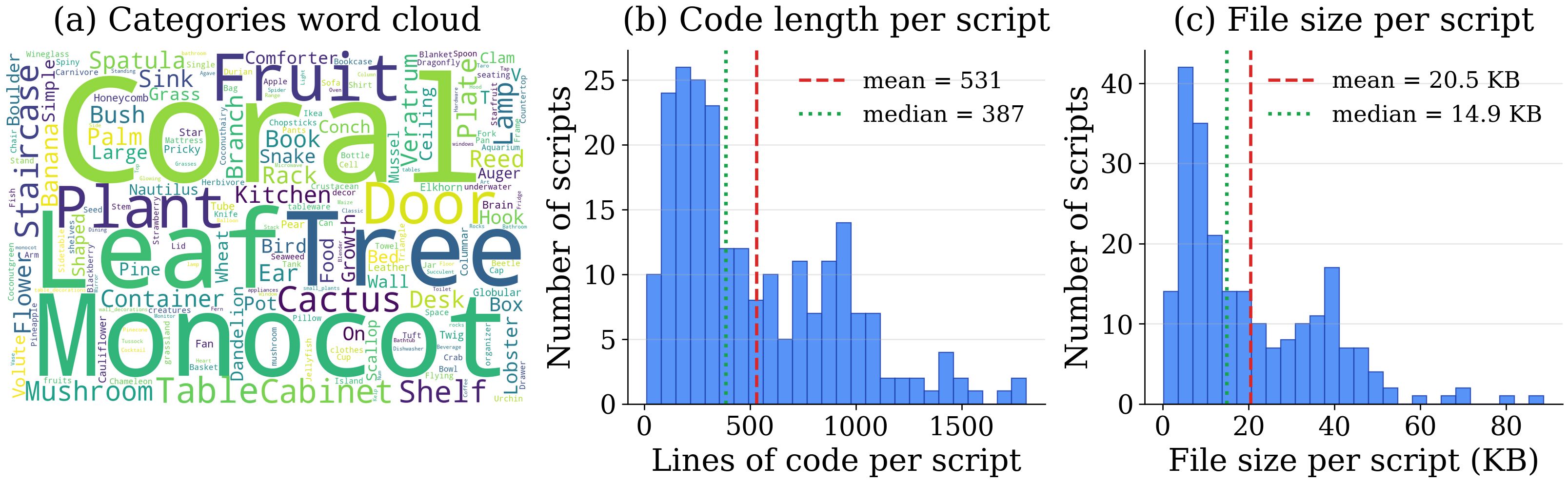}
  \caption{\textbf{\ourdata{} dataset statistics.} (a) A semantic word cloud illustrating the diversity of the 212 curated categories. (b) Distribution of code-line counts per script (mean $531$, median $387$), highlighting the script complexity. (c) Distribution of file sizes per script (mean $20.5$ KB, median $14.9$ KB), with a long right tail driven by intricate geometry-node definitions.}
  \label{fig:stats}
\end{figure}

The \ourdata{} benchmark includes a diverse taxonomy of \textbf{212} distinct asset categories from the Infinigen~\citep{raistrick2023infinigen}. 
As shown in Figure~\ref{fig:stats}(a), the semantic vocabulary is broad, encompassing organic entities (such as flora, fauna, and mollusks), manufactured objects (such as furniture and kitchenware), and architectural fragments. This level of coverage exceeds that of previous programmatic 3D benchmarks.
Figures~\ref{fig:stats}(b) and (c) present the distributions of code length and file size per script, both exhibiting strong right skew. The median script length is $387$ lines (mean $531$), with some scripts exceeding $1{,}000$ lines for complex geometry-node factories, including creature, tree, and cabinet variants. This complexity is intentional; whereas previous benchmarks typically assess simple primitive composition or voxel manipulation, \ourdata{} requires reasoning about 3D structure and newly introduced API functions.

\textbf{Curated High-quality Standalone 3D Code Data.}
Beyond the $212$-instance evaluation set, we also curate \textbf{3D Code Data} from the Infinigen~\citep{raistrick2023infinigen, raistrick2024infinigenindoors} simulator, a substantially larger corpus for supervised fine-tuning and procedural code research. 
Filtering the \textbf{$212$ random-seed-parameterized factories} underlying the $243$ full object factories yields \textbf{$12{,}963$ instances}: each is an \emph{(input prompt, standalone 3D code, 3D object)} triplet of a text description, $4$ canonical multi-view reference images ($45^{\circ}/135^{\circ}/225^{\circ}/315^{\circ}$), two Blender~5.0 Python scripts (a textured factory script and a geometry-only variant, resulting in \textbf{$\sim\!26$K code samples} in total), and the baked GLB ground-truth mesh, paired with three caption styles (object description, procedural-modeling instruction, factory-level specification). 
Every triplet has passed the agentic curation pipeline of Section~\ref{sec:pipeline1} including human-in-the-loop verification.

\subsection{Evaluation Protocol}
\label{sec:protocol}

Each policy is evaluated along two complementary axes: a quantitative metric suite that scores each generated mesh against the reference, and \textbf{3DCodeArena}, a public human-vote arena that ranks policies based on pairwise human preference.

\textbf{Quantitative metrics.}
For a given condition $c$, the policy $\pi$ is tasked with synthesizing a Blender~5.0 Python script $f_\pi$. We first evaluate the binary \emph{executability} indicator $\mathbb{I}[\mathcal{E}(f_\pi) \neq \emptyset]$, verifying that the script executes end-to-end and outputs a valid 3D mesh $M_\pi$. Conditioned on successful execution, we compute a suite of continuous mesh-grounded similarities $\mathcal{D}(M_\pi, M^{\star})$. Perceptual fidelity is assessed by rendering $M_\pi$ from four canonical viewpoints ($45^{\circ}$, $135^{\circ}$, $225^{\circ}$, $315^{\circ}$) under the standardized Cycles rig and computing SigLIP-2~\citep{tschannen2025siglip2} and DINOv3~\citep{simeoni2025dinov3} cosine similarities against the reference views. Structural and multi-modal alignments are evaluated by exporting $M_\pi$ to the GLB format and computing the Chamfer Distance~\citep{fan2017psgn}, alongside Uni3D, which computes 3D-3D and cross-modal (text-/image-to-3D) cosine similarities. 
Every quality metric $\mathcal{D}$ is reported under two paradigms: \emph{conditional} (averaged strictly over instances with $\mathbb{I}{=}1$, isolating geometric generation from code executability) and \emph{penalized} (assigned zero when $\mathbb{I}{=}0$, meaning the script fails to execute correctly in Blender). These metrics apply identically to the multi-turn ($T{>}1$) setting, where the final performance is measured on $M_\pi^{(T)}$ following iterative refinement.

\textbf{3DCodeArena.}
Beyond per-mesh metrics, we run a public arena: for each prompt, we precompute every viewable model pair; the website serves a random pair side by side, and human voters pick \emph{a}, \emph{b}, \emph{tie}, or \emph{both bad}; per-modality (text-to-3D and image-to-3D) Elo is kept on a separate scale to avoid conflating the distinct skill sets each track exercises. Bradley computes ratings--Terry MLE in log-strength space (LMArena convention: ties and ``both bad'' both contribute 0.5/0.5), recentered to a mean of $1000$ and converted to Elo points at $400/\ln 10$ per logit, with $1000$-resample bootstrap $95$\% confidence intervals.
At the time of writing, the platform hosts 12 frontier models and has collected approximately $3{,}100$ human votes.
\section{Experiments}
\label{sec:experiments}

We evaluate 12 frontier Vision-Language Models on \ourdata{}; Gemini~2.5 Pro and GPT~5.4 Nano were also tested but were excluded due to single-turn Executability below 10\%. To keep perceptual comparisons fair, every 3DCodeArena pair contains only successfully executed scripts; all metrics use only successfully rendered meshes, with reasoning budgets averaged where applicable. 
% The study then proceeds in three parts: Section~\ref{sec:t1results} correlates automated perception metrics with human preference Elo; Section~\ref{sec:ablations} ablates the thinking budget and multi-view input in the single-turn regime; and Section~\ref{sec:agentic} adds a multi-turn agentic loop over the same prompts.

\subsection{Correlation between Human Preference and Perception Metrics}
\label{sec:t1results}

\begin{figure}[t]
  \centering
  \includegraphics[width=1.0\linewidth]{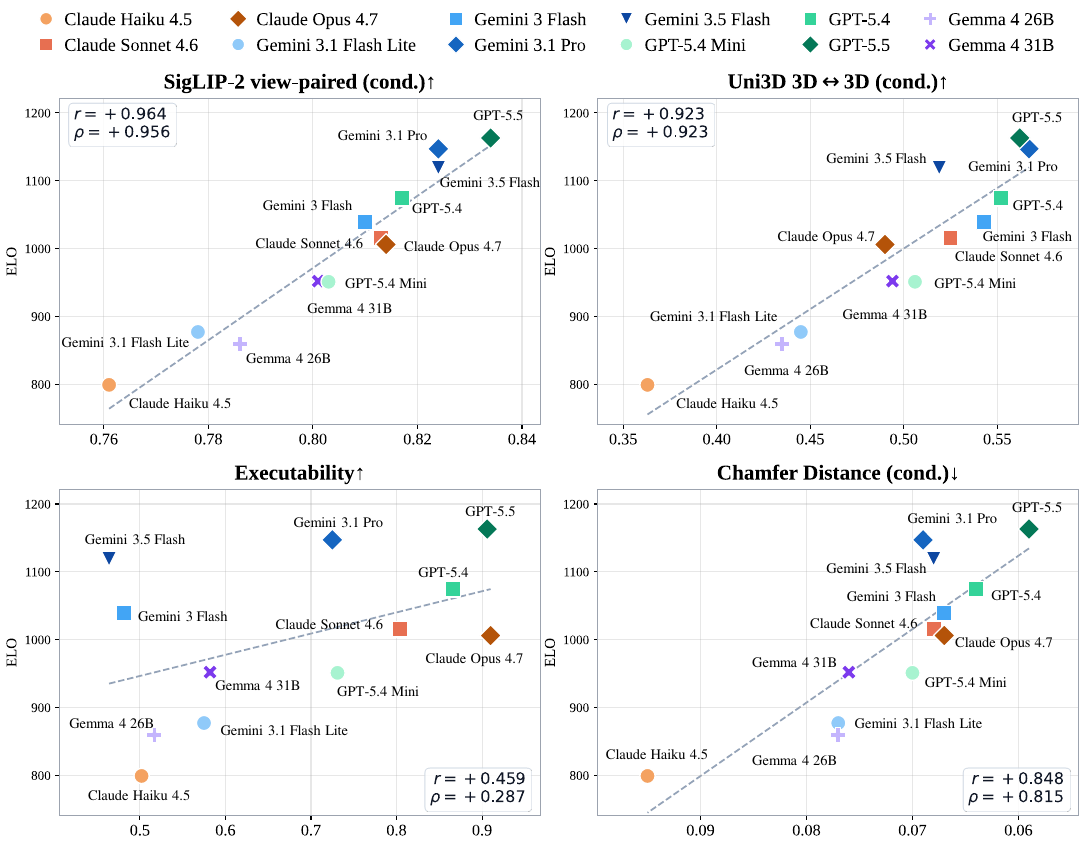}
  \caption{\textbf{Automated perception metrics vs.\ human preference.} For four representative per-model quality metrics, SigLIP-2 view-paired (cond.), Executability, Uni3D 3D$\leftrightarrow$3D (cond.), and Chamfer Distance (cond.), we plot the value (at each model's best thinking level, averaged across the text-to-3D and image-to-3D tracks) against the 3DCodeArena Elo on the $12$ evaluated VLMs and report Pearson $r$ and Spearman $\rho$. The Chamfer panel uses an inverted x-axis. 
  % so all panels read ``better metric $\rightarrow$ higher Elo''.
  }
  \label{fig:elo_vs_metrics}
\end{figure}

\begin{tcolorbox}[colback=blue!7,colframe=blue!45!black,boxrule=0.5pt,arc=2pt,left=8pt,right=8pt,top=5pt,bottom=5pt]
\textbf{Finding 1.} \emph{SigLIP-2 view similarity is one of the strongest predictors of human preference}.
\end{tcolorbox}

Figure~\ref{fig:elo_vs_metrics} illustrates the correlation between automated perception metrics and human preference Elo rankings collected from \emph{3DCodeArena}. The analysis demonstrates that automated multi-view similarity serves as a robust proxy for subjective human judgment. Specifically, SigLIP-2 view similarity is the strongest linear predictor of human preference (Pearson $r = 0.964$), while DINOv3 achieves the highest rank correlation (Spearman $\rho = 0.972$). These strong correlations across all 12 evaluated models validate the automated protocol, confirming that computationally scalable metrics such as SigLIP-2 and DINOv3 effectively capture the perceptual quality and structural integrity of generated 3D assets, thereby reliably substituting for costly human-in-the-loop annotations.

\subsection{Single-turn Ablation Studies on Thinking Budget and Multi-View Images}
\label{sec:ablations}

\begin{figure}[t]
  \centering
  \includegraphics[width=1.0\linewidth]{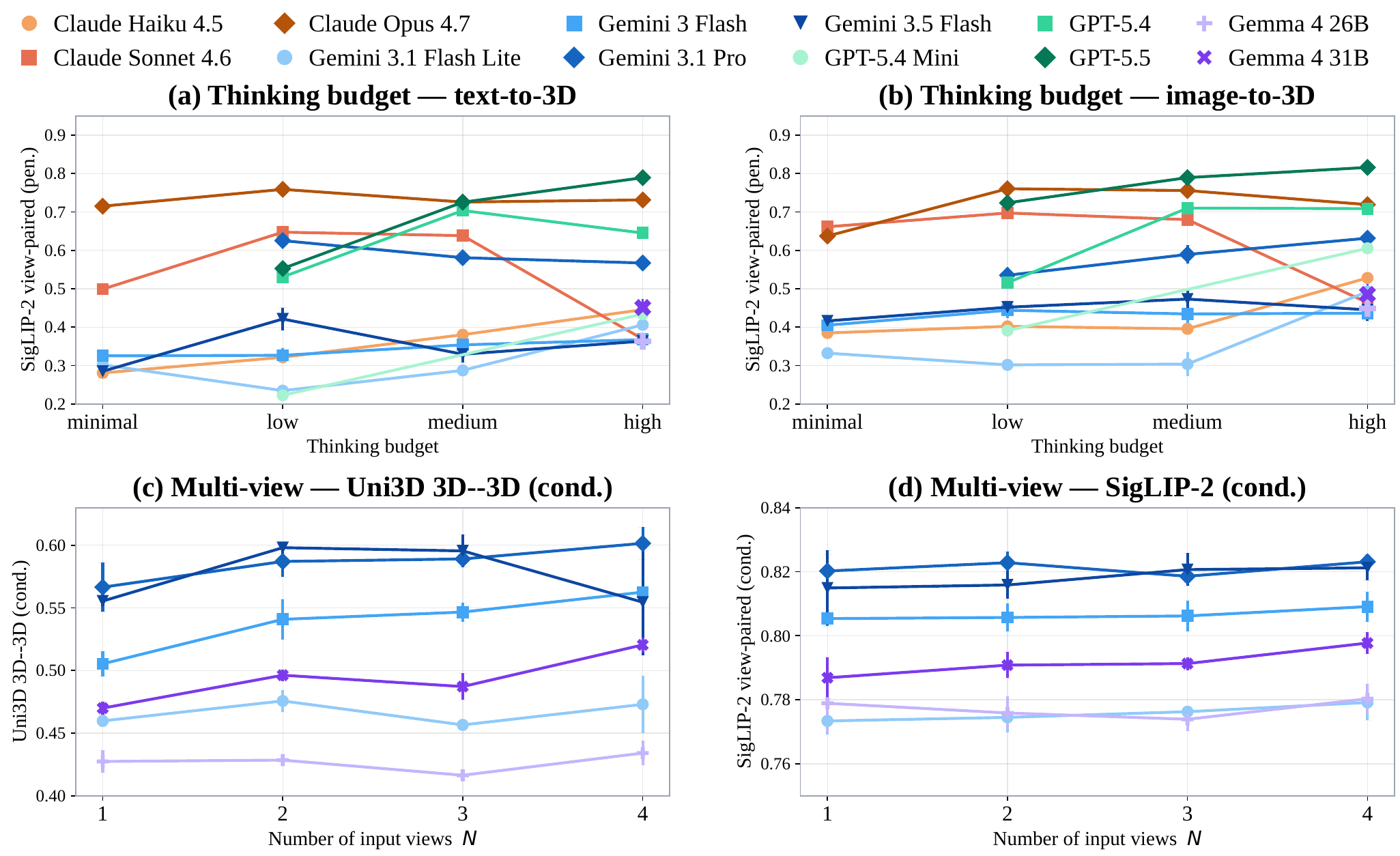}
  \caption{\textbf{Single-turn ablations.} Top row sweeps the thinking budget across all 12 VLMs on (a)~text-to-3D and (b)~image-to-3D, reported as SigLIP-2 view-paired (penalized) similarity. Bottom row sweeps the number of input reference views $N\in\{1,2,3,4\}$ on the image-to-3D track for the six backbones (four Gemini and two Gemma) we have multi-view runs for, reported as (c)~Uni3D 3D--3D (conditional) and (d)~SigLIP-2 (conditional), matching the successful-output quality convention in Table~\ref{tab:t1}. Solid lines with $\pm$std error bars use 3-seed means where multiple seeds are available; markers without error bars are single-seed runs.}
  \label{fig:abl-combined}
\end{figure}

\begin{tcolorbox}[colback=blue!7,colframe=blue!45!black,boxrule=0.5pt,arc=2pt,left=8pt,right=8pt,top=5pt,bottom=5pt]
\textbf{Finding 2.} \emph{Thinking budget helps lightweight reasoners but saturates early on frontier models.}
\end{tcolorbox}

\noindent\textbf{Thinking-level effort.}\quad
Figure~\ref{fig:abl-combined}(a,b) presents the effect of varying the thinking budget across all 12 backbones. Lightweight reasoners are considerably more sensitive to increased thinking budgets than heavier models: Gemini~3.1 Flash Lite gains approximately $19$ executability points from \texttt{minimal} to \texttt{high}, whereas Pro-class backbones (Gemini~3.1 Pro, Claude Opus~4.7, GPT-5.5) change by fewer than five points over the same range. 
The gain concentrates where baseline failures are dominated by Blender~5.0 API mismatches: extra reasoning tokens let lightweight backbones enumerate API alternatives and self-correct before emitting the script, whereas frontier models already encode the correct API and only marginally re-verify. 
Claude Opus~4.7 plateaus at \texttt{minimal}. Accordingly, \texttt{high} is adopted as the Flash-class default for the main results table (Table~\ref{tab:t1}).
Pairing the thinking level to capability (\texttt{high} for Flash/Haiku, \texttt{medium} for Pro/Sonnet/GPT-5.4, \texttt{minimal}--\texttt{low} for Opus/GPT-5.5) gives a $3$--$5\!\times$ cost reduction at comparable quality of the generated shapes.

\begin{tcolorbox}[colback=blue!7,colframe=blue!45!black,boxrule=0.5pt,arc=2pt,left=8pt,right=8pt,top=5pt,bottom=5pt]
\textbf{Finding 3.} \emph{Conditioned quality is largely insensitive to the input-view budget: extra views give at most modest Uni3D gains and no consistent SigLIP-2 gain over $N{=}1$.}
\end{tcolorbox}

\noindent\textbf{Multi-view image budget.}
Figure~\ref{fig:abl-combined}(c,d) sweeps the number of input views $N \in \{1,2,3,4\}$ on the image-to-3D track for the six backbones (four Gemini and two Gemma) for which we have multi-view runs.
Conditional SigLIP-2 view similarity (panel d) is nearly flat in $N$, varying by at most $0.012$ within each backbone; importantly, Gemini~3.5~Flash remains above both Gemma backbones under the same conditional convention as Table~\ref{tab:t1}.
Conditional Uni3D 3D--3D similarity (panel c) is likewise stable after scoring the same GLB-export path: per-backbone ranges stay within roughly $0.02$--$0.06$ across $N{=}1{-}4$.
Gemini~3.5~Flash peaks at $N{=}2{-}3$ ($0.60$) but returns to its $N{=}1$ level at $N{=}4$ ($0.56$), while Gemini~3.1~Pro changes from $0.57$ to $0.60$ over the sweep.
Table~\ref{tab:t1} uses $N{=}4$ to maintain cross-model comparability, while the conditional SigLIP ablation confirms that extra views provide no consistent view-similarity gains over $N{=}1{-}2$.
We use $N{=}4$ as the default for all the experiments, reserving multi-view inputs to test the ability of the VLMs on 3D spatial understanding.

\subsection{Evaluations on VLM Agents}
\label{sec:agentic}

Beyond the single-shot regime, we test whether an agentic workflow improves reliability or shape fidelity. The first is a \emph{multi-turn error-feedback retry}: a stateless, uniform loop that grants up to two additional attempts after any execution error in Table~\ref{tab:agentic}. The second hands \emph{full autonomy} to each model's native coding-agent harness, which freely writes, runs, and edits the script under a fixed wall-clock budget (Table~\ref{tab:agentic-harness}). 

\begin{table}[t]
\centering
\caption{\textbf{Multi-turn error-feedback on \ourdata{}.} For each instance whose single-turn render fails, we run up to two stateless retries that consume the previous code and the truncated Blender traceback. We report executability before/after the loop and the change in \emph{penalized mean} over all $212$ instances (failures contribute $0$; Chamfer uses a $1.5\!\times\!$max penalty), averaged across both tracks. Because the evaluation set is fixed, $\Delta$ cleanly captures overall benchmark lift without the set-shift artifact of conditional-mean comparison. \textbf{Bold} marks the executability ceiling.}
\label{tab:agentic}
\small
\setlength{\tabcolsep}{5pt}
\begin{tabular}{l cc ccc}
\toprule
 & \multicolumn{2}{c}{Executability\,$\uparrow$}
 & \multicolumn{3}{c}{$\Delta$\,penalized mean} \\
\cmidrule(lr){2-3} \cmidrule(lr){4-6}
Model
 & Single-turn
 & Multi-turn
 & SigLIP-2\,$\uparrow$
 & CD\,$\downarrow$
 & Uni3D 3D--3D\,$\uparrow$ \\
\midrule
Gemini~3 Flash         & 0.547 & 0.936          & $+$0.208 & $-$0.009 & $+$0.006 \\
Gemini~3.1 Flash Lite  & 0.580 & 0.929          & $+$0.164 & $+$0.000 & $+$0.000 \\
Gemini~3.1 Pro         & 0.698 & 0.993          & $+$0.145 & $-$0.193 & $+$0.130 \\
Gemini~3.5 Flash       & 0.479 & 0.946          & $+$0.201 & $+$0.036 & $+$0.212 \\
Gemma~4 26B            & 0.535 & 0.927          & $+$0.171 & $-$0.002 & $+$0.001 \\
Gemma~4 31B            & 0.554 & 0.976          & $+$0.204 & $+$0.001 & $-$0.001 \\
Claude Sonnet~4.6      & 0.804 & 0.993          & $+$0.068 & $-$0.130 & $+$0.058 \\
Claude Opus~4.7        & 0.910 & \textbf{1.000} & $+$0.033 & $-$0.084 & $+$0.056 \\
GPT-5.4 mini           & 0.731 & 0.995          & $+$0.110 & $-$0.237 & $+$0.124 \\
GPT-5.4                & 0.866 & \textbf{1.000} & $+$0.066 & $-$0.117 & $+$0.084 \\
GPT-5.5                & 0.906 & \textbf{1.000} & $+$0.040 & $-$0.129 & $+$0.086 \\
\midrule
\textbf{Aggregate}     & 0.692 & 0.972          & $+$0.128 & $-$0.079 & $+$0.069 \\
\bottomrule
\end{tabular}
\end{table}

\begin{tcolorbox}[colback=blue!7,colframe=blue!45!black,boxrule=0.5pt,arc=2pt,left=8pt,right=8pt,top=5pt,bottom=5pt]
\textbf{Finding 4.} \emph{Multi-turn error-feedback is effective and capacity-independent: it lifts Executability to near-ceiling and also improves overall benchmark quality on every backbone.}
\end{tcolorbox}

\noindent\textbf{Multi-turn error-feedback is a clean, capacity-independent win.}\quad
Aggregate executability across all $11{\times}2{=}22$ cells increases from $0.702$ (single-turn) to $0.974$ (multi-turn), representing a $+27.2$~pp improvement. \textbf{$8$ of $22$ cells reach the $1.000$ ceiling} (Claude~Opus~4.7, GPT-5.4, GPT-5.5 on both tracks, and Claude~Sonnet~4.6 / GPT-5.4-mini on image-to-3D). Beyond executability, the post-loop \emph{penalized mean} (failures contributing~$0$ on the fixed $212$-instance set) is positive across all $22$ cells on SigLIP-2 (aggregate $+0.128$) and on Uni3D~3D--3D for the high-capacity families (aggregate $+0.069$), and Chamfer Distance improves by $-0.079$ on aggregate; because the evaluation set is fixed, these deltas cleanly reflect overall benchmark lift rather than a set-shift artifact. Most of the improvement results from a single failure family: Blender~5.0 API mismatches whose fixes are localized and copy-pasteable, well within model competence once the traceback is visible (Appendix~\ref{app:multi-turn-debug}).

\begin{table}[t]
\centering
\caption{\textbf{Coding-agent harness on \ourdata{} text-to-3D.} Each backbone is wrapped in its native coding-agent harness (Gemini CLI for Gemini, Claude Code for Claude, Codex CLI for GPT-5.x, Antigravity CLI for Gemini 3.5 Flash), given the same task description as the single-turn baseline.
Each metric appears under two columns: \textbf{ST} (single-turn, no agent) vs.\ \textbf{Agent} (with the harness); shape metrics are conditional means over each side's own success set (every successfully rendered instance contributes to its column's average). \textbf{Bold} marks ceiling executability.}
\label{tab:agentic-harness}
\small
\setlength{\tabcolsep}{5.5pt}
\resizebox{\textwidth}{!}{%
\begin{tabular}{l l cc cc cc cc}
\toprule
 & & \multicolumn{2}{c}{Executability $\uparrow$}
   & \multicolumn{2}{c}{SigLIP-2 (cond.) $\uparrow$}
   & \multicolumn{2}{c}{Uni3D 3D--3D $\uparrow$}
   & \multicolumn{2}{c}{CD (cond.) $\downarrow$} \\
\cmidrule(lr){3-4} \cmidrule(lr){5-6} \cmidrule(lr){7-8} \cmidrule(lr){9-10}
Backbone & Harness & ST & Agent & ST & Agent & ST & Agent & ST & Agent \\
\midrule
Gemini 3 Flash             & Gemini CLI      & 0.608 & 0.995          & 0.175 & 0.170 & 0.495 & 0.517 & 0.071 & 0.072 \\
Gemini 3.1 Flash Lite      & Gemini CLI      & 0.608 & \textbf{1.000} & 0.155 & 0.121 & 0.412 & 0.366 & 0.078 & 0.087 \\
Gemini 3.1 Pro             & Gemini CLI      & 0.703 & 0.991          & 0.175 & 0.173 & 0.514 & 0.515 & 0.073 & 0.078 \\
Gemini 3.5 Flash           & Antigravity CLI     & 0.448 & 0.986          & 0.183 & 0.162 & 0.574 & 0.543 & 0.072 & 0.074 \\
\midrule
Claude Sonnet 4.6          & Claude Code     & 0.778 & 0.986          & 0.178 & 0.166 & 0.507 & 0.508 & 0.074 & 0.062 \\
Claude Opus 4.7            & Claude Code     & 0.887 & \textbf{1.000} & 0.179 & 0.175 & 0.465 & 0.533 & 0.068 & 0.071 \\
\midrule
GPT-5.4 mini               & Codex CLI       & 0.670 & \textbf{1.000} & 0.171 & 0.154 & 0.492 & 0.450 & 0.071 & 0.072 \\
GPT-5.4                    & Codex CLI       & 0.863 & \textbf{1.000} & 0.175 & 0.168 & 0.520 & 0.493 & 0.068 & 0.068 \\
GPT-5.5                    & Codex CLI       & 0.877 & 0.995          & 0.186 & 0.186 & 0.526 & 0.522 & 0.066 & 0.065 \\
\midrule
Average                    & ---             & 0.716 & 0.995          & 0.173 & 0.163 & 0.506 & 0.494 & 0.071 & 0.071 \\
\bottomrule
\end{tabular}%
}
\end{table}

\begin{tcolorbox}[colback=blue!7,colframe=blue!45!black,boxrule=0.5pt,arc=2pt,left=8pt,right=8pt,top=5pt,bottom=5pt]
\textbf{Finding 5.} \emph{Harnesses can lift Executability but, scored on the same instance subset, produce shape fidelity indistinguishable from a single prompt.}
\end{tcolorbox}

\noindent\textbf{Coding-agent harnesses lift Executability further but do not improve conditional shape quality.}\quad
The retry loop described above is stateless and uniform across backbones. To determine whether \emph{full agent autonomy} provides additional benefits beyond error-feedback fixes, all eight in-budget backbones were re-run within their native coding-agent harness: Gemini CLI for the three Gemini variants, Claude Code for Sonnet and Opus, and Codex CLI for the GPT-5.x family. Within a wall-clock time budget of $600$--$900$\,s, the agent receives the same task description, autonomously writes the script, invokes Blender~5.0, edits the file, and iterates until a mesh is produced, or the budget expires. Aggregated across the eight backbones (Table~\ref{tab:agentic-harness}), the harness increases executability from $0.747$ to $0.973$, a $+22.6$\,pp gain comparable to the stateless retry of Finding~4, with three of eight backbones reaching the $1.000$ ceiling and a further two at $\geq 0.99$. When restricted to the ST-success $\cap$ Agent-success \emph{intersection} so both columns score on the same instance subset, conditional SigLIP-2 view similarity changes by only $-0.010$ on average, conditional Chamfer Distance by only $+0.001$, and conditional Uni3D~3D--3D similarity (the metric most strongly tracking human preference per Finding~1) by only $-0.003$. The harness addresses simple API usage errors. However, it does not yield semantically richer or more shape-accurate geometry once a script compiles.

\subsection{Qualitative Comparison}
\label{sec:qualitative}

\begin{figure}[tbp]
  \centering
  \includegraphics[width=1.0\linewidth]{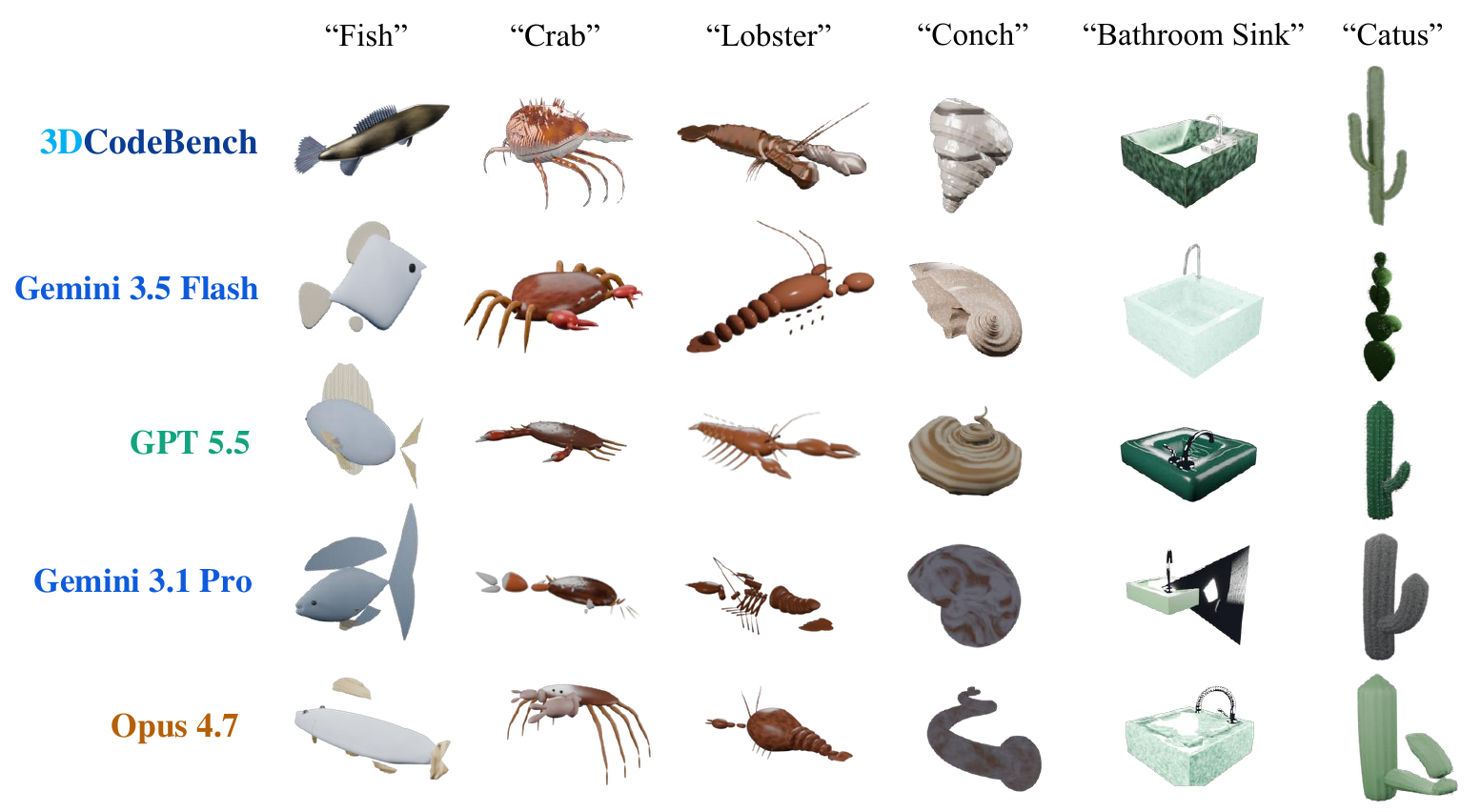}
  \caption{\textbf{Qualitative comparison} of Gemini~3.1~Pro, Claude Opus~4.7, and GPT-5.5 against the 3DCodeBench reference on six prompts. Every mesh is the render of each model's procedural output under the agentic workflow.}
  \label{fig:comparison}
\end{figure}

Figure~\ref{fig:comparison} presents a qualitative comparison across six held-out prompts (e.g., ``Fish'', ``Lobster'', ``Bathroom Sink'') evaluating three frontier models against the 3DCodeBench reference using the corresponding agent harness. Each mesh shows the rendered output of the model's procedural pipeline. While these models capture basic silhouettes, they often struggle with structural integrity, frequently degenerating into disconnected geometric fragments (e.g., Gemini~3.1~Pro) or simplistic, floating primitives (e.g., Opus~4.7).

\section{Conclusion and Future Work}
\label{sec:conclusion}
\textbf{Conclusion.} In this paper, we introduced \ourdata{}, a benchmark for evaluating vision-language model (VLM) agents in procedural 3D modeling. Using a novel agentic curation pipeline based on the Infinigen simulator, we constructed a diverse dataset of 212 object categories paired with executable code. Extensive evaluations of 12 frontier VLMs, which combine automated metrics with \emph{3DCodeArena} human preferences, highlight a critical capability gap: while models can produce executable code, they struggle with complex geometric reasoning and physical plausibility. Crucially, test-time scaling and multi-turn agentic refinement effectively mitigate these shortcomings. We also establish SigLIP-2 view similarity as a robust automated proxy for human judgment.

\textbf{Future work.} In the future, we plan to extend \ourdata{} to multi-asset scene composition and evaluate cross-platform versatility (e.g., SideFX Houdini or Unreal Engine) to disentangle API memorization from generalized procedural modeling capabilities. Furthermore, scaling our curation pipeline could yield larger-scale datasets for pre-training next-generation 3D-aware VLMs.
Ultimately, \ourdata{} and 3DCodeArena establish a foundational framework for advancing autonomous agents capable of generating high-quality 3D shapes.

\bibliography{refs}

% APPENDIX
\appendix
% =========================================================================
% appendix.tex -- appendix content for the \ourdata{} submission.
% Included by neurips_2026.tex after \bibliography{refs} via \input.
% The main file is responsible for issuing \appendix before this file is
% input so that sections are numbered A, B, C, ...
% =========================================================================

% \msd defined in the main preamble (neurips_2026.tex).

\clearpage

\renewcommand{\thetable}{\thesection.\arabic{table}}
\renewcommand{\thefigure}{\thesection.\arabic{figure}}
\counterwithin{table}{section}
\counterwithin{figure}{section}

\section{Appendix}
\label{app:results}

\subsection{Cost--Quality Pareto Frontier}
\label{app:cost-elo}

\begin{figure}[htbp]
\centering
\includegraphics[width=0.99\linewidth]{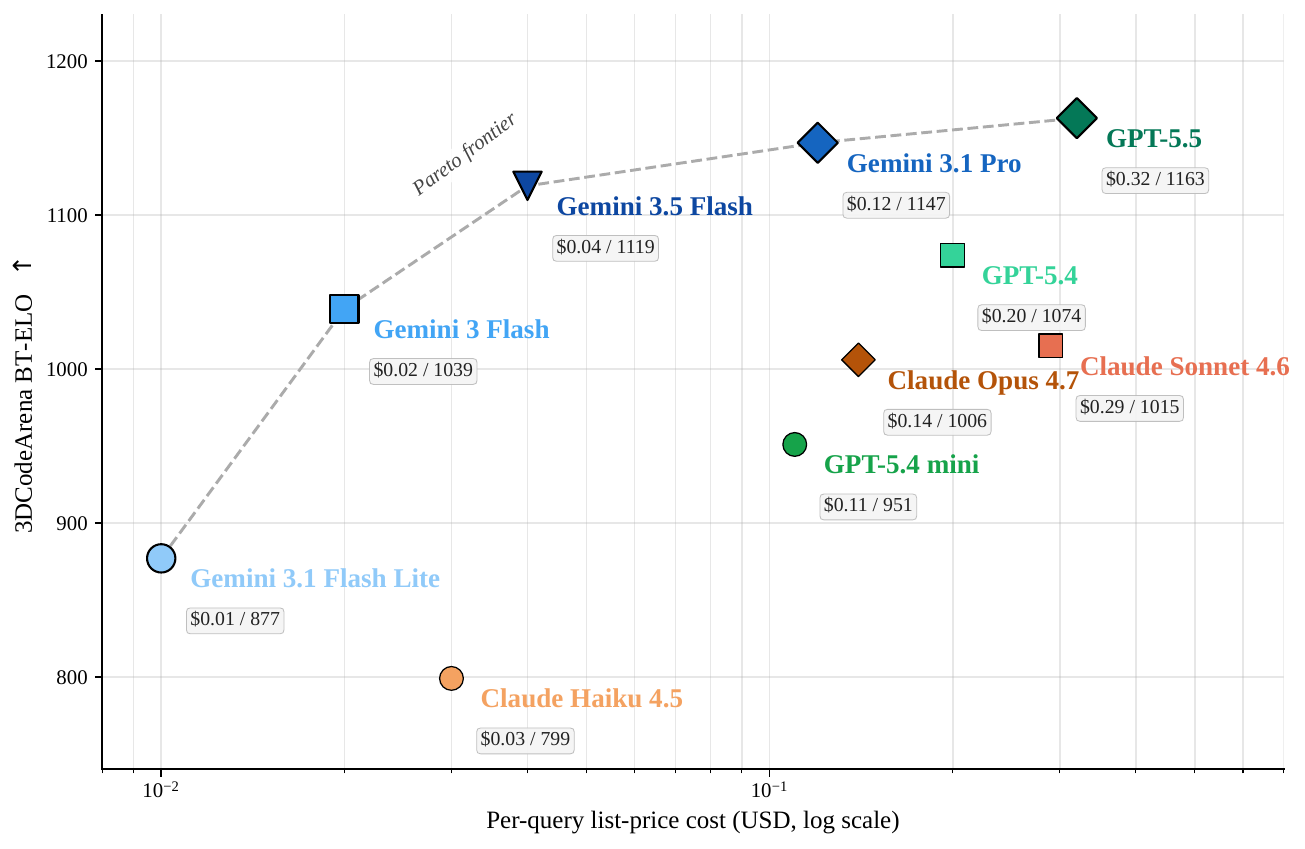}
\caption{\textbf{Cost versus human-preference Elo across the $10$ paid frontier VLMs.} The $x$-axis is per-query list-price cost (USD, log scale) and the $y$-axis is 3DCodeArena Bradley--Terry Elo (higher is better; $3{,}098$ votes). Each point is labeled with its (cost~/~Elo) and the dashed line is the Pareto frontier. Four of the five frontier points are Gemini models, and GPT-5.5 is the only non-Gemini point on it.}
\label{fig:elo_vs_cost}
\end{figure}

Figure~\ref{fig:elo_vs_cost} plots each model's 3DCodeArena Bradley--Terry Elo against its per-query cost. We define cost as the dollar price of a single generation under each provider's published API rates. The two free Gemma backbones are omitted so that the cost axis remains meaningful, leaving the $10$ paid frontier VLMs. The dashed line marks the Pareto frontier, that is, the cheapest model that reaches each Elo level as cost increases.

The plot shows three patterns. First, the Gemini models make up four of the five points on the frontier: Gemini~3.1~Flash~Lite (\$0.01, $877$), Gemini~3~Flash (\$0.02, $1039$), Gemini~3.5~Flash (\$0.04, $1119$), and Gemini~3.1~Pro (\$0.12, $1147$); GPT-5.5 (\$0.32, $1163$) is the only non-Gemini point on the frontier and sits at the top. Second, Elo rises quickly at low cost and then flattens, so Gemini~3.5~Flash comes within about $44$ Elo of the best model at roughly one-eighth of its cost. Third, the Claude models fall below the frontier: Claude~Opus~4.7 (\$0.14, $1006$) and Claude~Sonnet~4.6 (\$0.29, $1015$) score about $130$ to $160$ Elo lower than Gemini models at similar or lower cost. GPT-5.4~mini and Claude~Haiku~4.5 are cheap but score low, and GPT-5.4 (\$0.20, $1074$) also stays below the frontier at its price.

\subsection{Per-Model Main-Results Table}
\label{app:full-main-results}

Table~\ref{tab:t1} reports the per-model numbers behind Figure~\ref{fig:elo_vs_metrics}: for each of the $12$ evaluated VLMs, we select the single thinking-effort level that maximizes SigLIP-2 (best level) and report all metrics at that level, averaged across both the text-to-3D and image-to-3D tracks. Claude and GPT have only one inference-time setting (their base run), so their best level equals the base run; Gemma has only one API-permitted level (\texttt{high}). Only Gemini models have multiple thinking levels to select from.

% The model rankings are virtually the same using SigLIP or Dinov3 similarity scores (where the rankings differ it is by models being swapped by epsilon difference in score).   This strong correlation is interesting because these models are trained quite differently (siglip is text-aligned like CLIP, and dino is self-supervised).   Same observation for the image-to-3d table.
\begin{table}[t!]
\centering
\caption{Main results on \ourdata{} (212 categories). \textbf{Each model is evaluated at its best thinking level} --- the single thinking-effort setting that maximizes SigLIP-2 --- averaged across both the text-to-3D and image-to-3D tracks (conditional mean). Claude and GPT have only one inference-time setting (their base run); Gemma has only one API-permitted level (\texttt{high}); for these families best-level equals the base run. Single-shot and thinking-average breakdowns are in Appendix~\ref{app:agg-comparison}.
\emph{Exec.}\,$\uparrow$: Blender~5.0 pass rate.
\emph{Image-grounded}: SigLIP-2 / DINOv3 cosine between rendered and reference views (conditional).
\emph{3D-shape}: Chamfer, Uni3D 3D--3D paired, and Uni3D cross-modal cosine on the exported GLB.
\emph{ELO}: combined Bradley--Terry Elo on 3DCodeArena.
\emph{Per-query cost}: mean output tokens, wall-clock time, throughput, and list-price spend.}
\label{tab:t1}
\scriptsize
\setlength{\tabcolsep}{3pt}
\begin{tabular}{l c cc ccc c rrrr}
\toprule
 & & \multicolumn{2}{c}{Image-grounded\,$\uparrow$}
 & \multicolumn{3}{c}{3D-shape}
 & & \multicolumn{4}{c}{Per-query cost} \\
\cmidrule(lr){3-4} \cmidrule(lr){5-7} \cmidrule(lr){9-12}
Model
 & Exec.\,$\uparrow$
 & SigLIP-2 & DINOv3
 & Chamfer\,$\downarrow$ & Uni3D\,$\uparrow$ & \makecell{Uni3D\\t/i--3D\,$\uparrow$}
 & ELO\,$\uparrow$
 & Tok.
 & Time (s)
 & Tok/s
 & Cost (\$) \\
\midrule
Gemini~3 Flash         & 0.481 & 0.810 & 0.528 & 0.067 & 0.543 & 0.277 & 1{,}039 &  2{,}647 &  34.0 &  78 & 0.02 \\
Gemini~3.1 Flash Lite  & 0.575 & 0.778 & 0.496 & 0.077 & 0.445 & 0.246 &   877 &    875 &  36.7 &  24 & 0.01 \\
Gemini~3.1 Pro         & 0.725 & 0.824 & 0.569 & 0.069 & \textbf{0.567} & 0.284 & 1{,}147 &  2{,}030 & 162.9 &  12 & 0.19 \\
Gemini~3.5 Flash       & 0.464 & 0.824 & 0.563 & 0.068 & 0.519 & 0.266 & 1{,}119 &  3{,}590 &  93.3 &  38 & 0.04 \\
Gemma~4 26B            & 0.517 & 0.786 & 0.483 & 0.077 & 0.435 & 0.248 &   859 &  2{,}678 & 113.7 &  24 & free \\
Gemma~4 31B            & 0.582 & 0.801 & 0.518 & 0.076 & 0.494 & 0.261 &   952 &  1{,}732 & 119.1 &  15 & free \\
Claude Haiku~4.5       & 0.502 & 0.761 & 0.413 & 0.095 & 0.363 & 0.219 &   799 &  3{,}770 &  23.3 & 162 & 0.02 \\
Claude Sonnet~4.6      & 0.804 & 0.813 & 0.551 & 0.068 & 0.525 & 0.277 & 1{,}015 & \textbf{16{,}508} & 200.3 &  82 & 0.26 \\
Claude Opus~4.7        & \textbf{0.910} & 0.814 & 0.545 & 0.067 & 0.490 & 0.268 & 1{,}006 &  2{,}363 &  30.0 &  79 & 0.08 \\
GPT-5.4 mini           & 0.731 & 0.803 & 0.526 & 0.070 & 0.506 & 0.275 &   951 &  2{,}402 & 155.8 &  15 & 0.10 \\
GPT-5.4                & 0.866 & 0.817 & 0.560 & 0.064 & 0.552 & \textbf{0.285} & 1{,}074 &  2{,}725 & 168.0 &  16 & 0.18 \\
GPT-5.5                & 0.906 & \textbf{0.834} & \textbf{0.576} & \textbf{0.059} & 0.562 & 0.284 & \textbf{1{,}163} &  3{,}748 & 160.1 &  23 & 0.28 \\
\bottomrule
\end{tabular}
\end{table}

\subsection{Elo vs.\ Metrics under Single-Shot Run}
\label{app:elo-vs-metrics-bothtrack}

\begin{figure}[htbp]
\centering
\includegraphics[width=\linewidth]{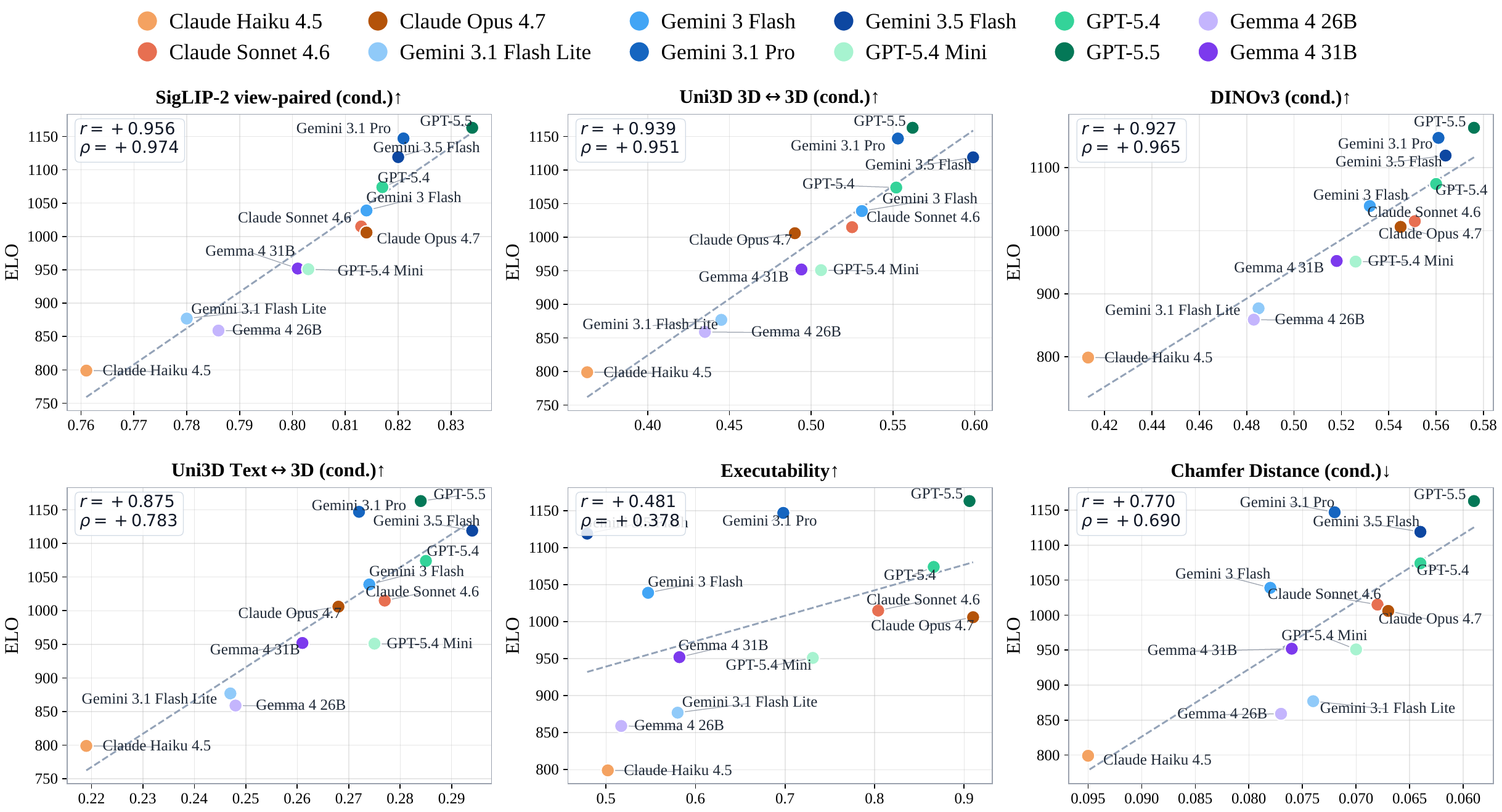}
\caption{\textbf{3DCodeArena Elo vs.\ each automated quality metric, single-shot run.} Same six-panel layout as Figure~\ref{fig:elo_vs_metrics} but using each model's single-shot base run (rather than best thinking level) averaged across both tracks. Per-panel Pearson $r$ and Spearman $\rho$ are reported in the upper-left; the Chamfer panel uses an inverted x-axis (smaller is better) so all panels read left-to-right as ``better metric $\to$ higher Elo''.}
\label{fig:elo_vs_metrics_bothtrack}
\end{figure}

The main-paper Figure~\ref{fig:elo_vs_metrics} uses the \emph{best-level} aggregation. Figure~\ref{fig:elo_vs_metrics_bothtrack} repeats the same six-panel analysis under the \emph{single-shot} convention --- each model's base run (one model call per instance at a fixed thinking setting), averaged across both tracks --- as a robustness check. The rankings are essentially unchanged: image-grounded similarity and 3D-shape metrics remain strong monotone predictors of human preference, with the cross-modal and 3D--3D Uni3D panels the strongest. Executability is the weakest correlate, consistent with our finding that physical plausibility, not mere code execution, drives human preference.

\begin{figure}[htbp]
\centering
\includegraphics[width=\linewidth]{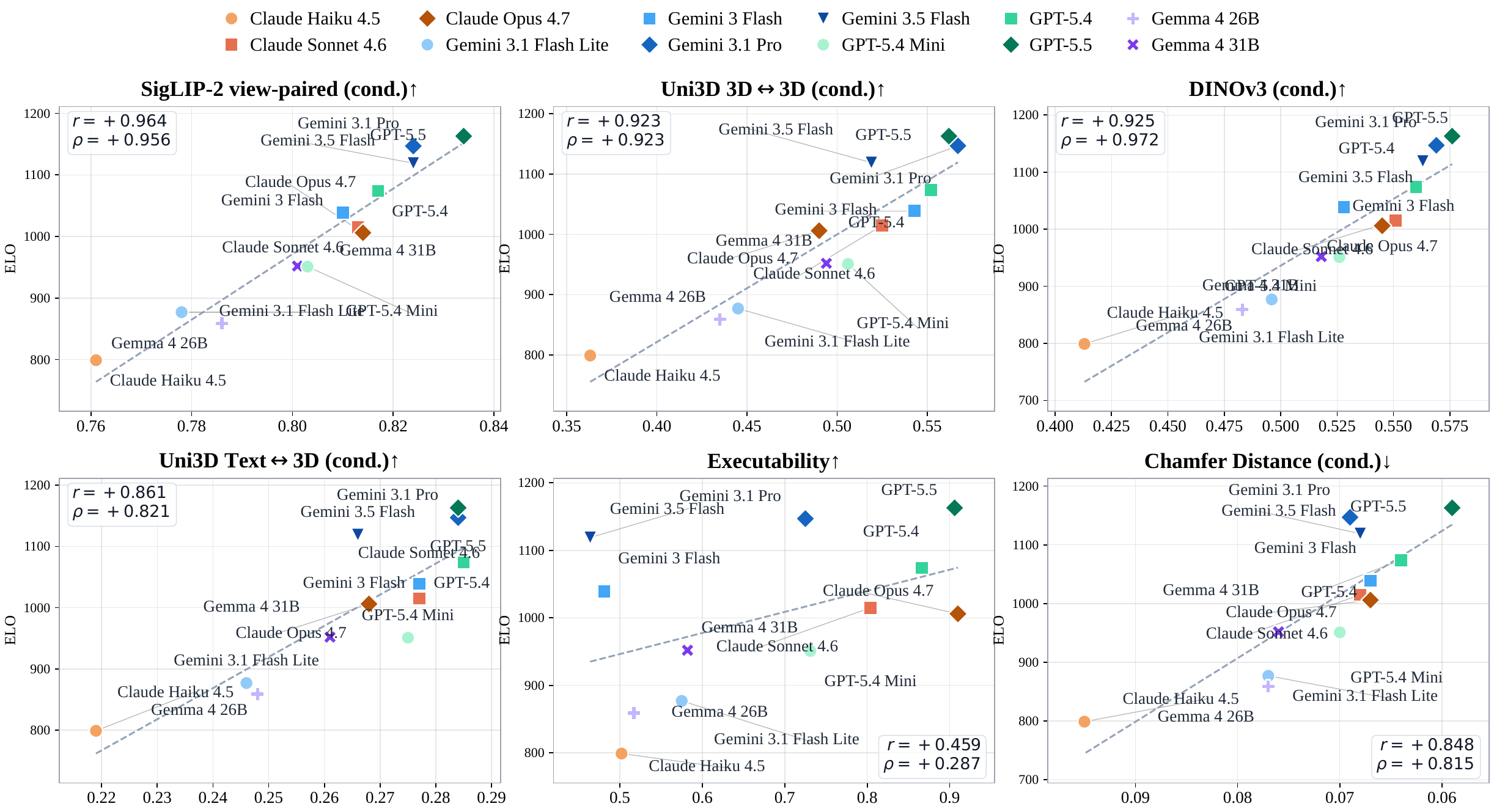}
\caption{\textbf{3DCodeArena Elo vs.\ all six automated quality metrics (best-level aggregation).} The full six-panel version of the main-paper Figure~\ref{fig:elo_vs_metrics}, which shows only four panels for space. Same per-model palette; per-panel Pearson $r$ and Spearman $\rho$ in the upper-left; the Chamfer panel uses an inverted x-axis so all panels read ``better metric $\to$ higher Elo''.}
\label{fig:elo_vs_metrics_full}
\end{figure}

\subsection{Aggregation Comparison: Single-Shot vs.\ Thinking-Average}
\label{app:agg-comparison}

The headline Table~\ref{tab:t1} uses best-level aggregation. For completeness, Tables~\ref{tab:singleshot-combined} and~\ref{tab:thinking-avg-combined} report the same $12$ models under two alternative aggregation methods: single-shot (one base run per model, fixed thinking level) and thinking-level average (all available levels averaged, with Claude/GPT falling back to their base run). The three aggregations produce very similar model rankings; the primary differences are on Gemini models, where the best level can differ from the all-level mean by up to $0.08$ on Uni3D (e.g.\ Gemini~3.5~Flash: best-level $0.519$ vs.\ single-shot $0.599$, the latter benefiting from its base run happening to land on a strong Uni3D level).

% Single-shot base run, both-track average (text-to-3D + image-to-3D).
% Claude/GPT use their base run; Gemma (no separate base) uses its one API-permitted level.
\begin{table}[htbp]
\centering
\caption{\textbf{Single-shot run} (one model call per instance, no thinking-level search).
Per-model values are averaged across both the text-to-3D and image-to-3D tracks (conditional mean).
Open-weight Gemma has no separate base run and is evaluated at its API-default thinking level.}
\label{tab:singleshot-combined}
\scriptsize
\setlength{\tabcolsep}{4pt}
\begin{tabular}{l c cc ccc c}
\toprule
 & & \multicolumn{2}{c}{Image-grounded\,$\uparrow$}
 & \multicolumn{3}{c}{3D-shape}
 & \\
\cmidrule(lr){3-4} \cmidrule(lr){5-7}
Model
 & Exec.\,$\uparrow$
 & SigLIP-2 & DINOv3
 & Chamfer\,$\downarrow$ & Uni3D\,$\uparrow$ & \makecell{Uni3D\\t/i--3D\,$\uparrow$}
 & ELO\,$\uparrow$ \\
\midrule
Gemini~3 Flash         & 0.547 & 0.814 & 0.532 & 0.078 & 0.531 & 0.274 & 1{,}039 \\
Gemini~3.1 Flash Lite  & 0.580 & 0.780 & 0.485 & 0.074 & 0.445 & 0.247 &   877 \\
Gemini~3.1 Pro         & 0.698 & 0.821 & 0.561 & 0.072 & 0.553 & 0.272 & 1{,}147 \\
Gemini~3.5 Flash       & 0.479 & 0.820 & 0.564 & 0.064 & \textbf{0.599} & \textbf{0.294} & 1{,}119 \\
Gemma~4 26B            & 0.517 & 0.786 & 0.483 & 0.077 & 0.435 & 0.248 &   859 \\
Gemma~4 31B            & 0.582 & 0.801 & 0.518 & 0.076 & 0.494 & 0.261 &   952 \\
Claude Haiku~4.5       & 0.502 & 0.761 & 0.413 & 0.095 & 0.363 & 0.219 &   799 \\
Claude Sonnet~4.6      & 0.804 & 0.813 & 0.551 & 0.068 & 0.525 & 0.277 & 1{,}015 \\
Claude Opus~4.7        & \textbf{0.910} & 0.814 & 0.545 & 0.067 & 0.490 & 0.268 & 1{,}006 \\
GPT-5.4 mini           & 0.731 & 0.803 & 0.526 & 0.070 & 0.506 & 0.275 &   951 \\
GPT-5.4                & 0.866 & 0.817 & 0.560 & 0.064 & 0.552 & 0.285 & 1{,}074 \\
GPT-5.5                & 0.906 & \textbf{0.834} & \textbf{0.576} & \textbf{0.059} & 0.562 & 0.284 & \textbf{1{,}163} \\
\bottomrule
\end{tabular}
\end{table}

% Thinking-average, both-track average (text-to-3D + image-to-3D).
% Gemini/Gemma: average across all thinking-effort levels × 3 seeds.
% Claude/GPT: fall back to their single-shot base run (no thinking ablation).
\begin{table}[htbp]
\centering
\caption{\textbf{Thinking-level average} (all available thinking-effort levels averaged per model).
Gemini and Gemma models are averaged across four thinking levels (\texttt{minimal}/\texttt{low}/\texttt{medium}/\texttt{high}) at three seeds each; Claude and GPT have no thinking-ablation runs and fall back to their single-shot base run.
Values are conditional means averaged across both the text-to-3D and image-to-3D tracks.}
\label{tab:thinking-avg-combined}
\scriptsize
\setlength{\tabcolsep}{4pt}
\begin{tabular}{l c cc ccc c}
\toprule
 & & \multicolumn{2}{c}{Image-grounded\,$\uparrow$}
 & \multicolumn{3}{c}{3D-shape}
 & \\
\cmidrule(lr){3-4} \cmidrule(lr){5-7}
Model
 & Exec.\,$\uparrow$
 & SigLIP-2 & DINOv3
 & Chamfer\,$\downarrow$ & Uni3D\,$\uparrow$ & \makecell{Uni3D\\t/i--3D\,$\uparrow$}
 & ELO\,$\uparrow$ \\
\midrule
Gemini~3 Flash         & 0.478 & 0.806 & 0.525 & 0.074 & 0.529 & 0.273 & 1{,}039 \\
Gemini~3.1 Flash Lite  & 0.430 & 0.770 & 0.456 & 0.084 & 0.422 & 0.239 &   877 \\
Gemini~3.1 Pro         & 0.713 & 0.816 & 0.558 & 0.073 & 0.553 & 0.278 & 1{,}147 \\
Gemini~3.5 Flash       & 0.454 & 0.816 & 0.558 & 0.072 & 0.470 & 0.250 & 1{,}119 \\
Gemma~4 26B            & 0.517 & 0.786 & 0.483 & 0.077 & 0.435 & 0.248 &   859 \\
Gemma~4 31B            & 0.582 & 0.801 & 0.518 & 0.076 & 0.494 & 0.261 &   952 \\
Claude Haiku~4.5       & 0.502 & 0.761 & 0.413 & 0.095 & 0.363 & 0.219 &   799 \\
Claude Sonnet~4.6      & 0.804 & 0.813 & 0.551 & 0.068 & 0.525 & 0.277 & 1{,}015 \\
Claude Opus~4.7        & \textbf{0.910} & 0.814 & 0.545 & 0.067 & 0.490 & 0.268 & 1{,}006 \\
GPT-5.4 mini           & 0.731 & 0.803 & 0.526 & 0.070 & 0.506 & 0.275 &   951 \\
GPT-5.4                & 0.866 & 0.817 & 0.560 & 0.064 & 0.552 & \textbf{0.285} & 1{,}074 \\
GPT-5.5                & 0.906 & \textbf{0.834} & \textbf{0.576} & \textbf{0.059} & \textbf{0.562} & 0.284 & \textbf{1{,}163} \\
\bottomrule
\end{tabular}
\end{table}

\FloatBarrier

\subsection{Models Evaluated}
\label{app:models}

We benchmark \textbf{twelve} frontier vision--language models from four providers across three compute tiers: (i)~\emph{lightweight} reasoners optimized for low-latency completion (Claude~Haiku~4.5, Gemini~3.1~Flash~Lite~\citep{gemini3flashlite_blog}, GPT-5.4-mini, Gemma~4~26B and~31B~\citep{gemma4_modelcard}); (ii)~\emph{mid-tier} reasoners (Claude~Sonnet~4.6, Gemini~3 Flash, Gemini~3.5~Flash, GPT-5.4); and (iii)~\emph{frontier} reasoners (Claude~Opus~4.7, Gemini~3.1 Pro~\citep{gemini31pro_blog}, GPT-5.5). Two further models --- the prior-generation Gemini~2.5 Pro and GPT-5.4 Nano --- are run for the inclusion-threshold analysis below but excluded from the headline tables.

\textbf{User message and decoding parameters.}
All models are queried in a zero-shot setting using the task-specific system prompt (Appendix~\ref{app:inference}). The user message is the per-instance \texttt{prompt\_description.txt} on the text-to-3D track and the four canonical reference views (\texttt{Image\_\{005, 015, 025, 035\}.png}) on the image-to-3D track. Decoding parameters are held fixed across providers where the API exposes them: \texttt{temperature}~$=0.7$, a $65{,}536$-token output cap, and a defensive Markdown-fence stripper that drops any stray triple-backtick lines before the response is saved as a \texttt{.py} file and executed under Blender~5.0.

\textbf{Per-provider thinking-budget mapping.}
The four providers expose reasoning-budget control via different APIs; Table~\ref{tab:t1} reports each model at its best thinking level (the one maximizing SigLIP-2), so that it reflects each model's peak capability; alternative aggregations (single-shot base run and thinking-level average) are in Appendix~\ref{app:agg-comparison}.
\begin{itemize}[leftmargin=1.5em,itemsep=2pt,topsep=2pt]
    \item \emph{Gemini~3 family.} The official \texttt{thinking\_level} enum (\texttt{minimal}/\texttt{low}/\texttt{medium}/\texttt{high}) is exposed via \texttt{GenerateContentConfig}. We sweep all four levels at three seeds $\{0, 1, 2\}$ on Flash and Flash~Lite. Pro does not accept \texttt{minimal} via the API, so its lightest probed tier is \texttt{low} ($1.8$K average thinking tokens).
    \item \emph{Gemma~4 (26B / 31B).} The Gemini API does not expose a tunable \texttt{thinking\_level} for Gemma --- \texttt{minimal}/\texttt{low}/\texttt{medium} are rejected with \texttt{INVALID\_ARGUMENT}, and the open-weight HuggingFace release exposes only a binary \texttt{enable\_thinking} flag. We therefore report Gemma at the API-allowed \texttt{high} equivalent only.
    \item \emph{Claude~4 family.} The unified \texttt{thinking} enum is mapped to native \texttt{\{"type":\allowbreak{}"enabled",\allowbreak{}"budget\_tokens":\allowbreak{}N\}} with $N \in \{0, 4{\rm K}, 16{\rm K}, 32{\rm K}\}$ for \texttt{minimal}/\texttt{low}/\texttt{medium}/\texttt{high}, and an additional $N=64{\rm K}$ at \texttt{xhigh}. The Anthropic API does not accept a sampling \texttt{seed} when adaptive thinking is enabled, so Claude rows are single-seed.
    \item \emph{GPT-5 family.} OpenAI's \texttt{reasoning.effort} parameter (\texttt{minimal}/\texttt{low}/\texttt{medium}/\texttt{high}/\texttt{xhigh}) is passed straight through. We use each model's empirical peak for the headline table --- GPT-5.4-mini and GPT-5.4 at \texttt{medium}, GPT-5.5 at \texttt{high} --- and run three seeds per cell where available.
\end{itemize}

\textbf{Excluded models: prior-generation reasoners.}
Two models fall below our $10\%$ single-shot text-to-3D executability floor and are dropped from the main analysis: \textbf{Gemini~2.5 Pro} achieves $0.071$ executability on text-to-3D and $0.217$ on image-to-3D, and \textbf{GPT-5.4 Nano} achieves $0.061$ and $0.165$ respectively. Across both excluded backbones, $\sim\!85\%$ of failures are Blender~$4.x \to 5.0$ API drift errors --- e.g.\ \texttt{KeyError "Specular"} on the removed BSDF socket, \texttt{AttributeError 'Mesh'.use\_auto\_smooth}, \texttt{enum "SUBSURFACE" not found in ObjectModifiers.new}, and \texttt{TypeError create\_cone keyword "diameter1" is invalid}. The models are therefore not modeling-capacity-limited but knowledge-cutoff-limited: their training data predates Blender~5.0's API renames, and they cannot recover from the cascade of mismatches. The $10\%$ floor is chosen so that the conditional ($\mathbb{I}{=}1$) means in Table~\ref{tab:t1} remain reliable; below it, the rendered-only sample size is too small for stable per-metric estimates, and zeros dominate the penalized means. Multi-turn error feedback substantially closes this knowledge gap for the included backbones (Section~\ref{app:multi-turn-debug}); we leave its application to the excluded models as an exercise for future evaluators of older-generation reasoners.

\textbf{Multi-turn and agentic-harness protocols.}
The multi-turn loop of Section~\ref{sec:agentic} runs up to two additional \emph{stateless} retries per failed instance (\texttt{ERR\_EXEC} / \texttt{ERR\_NO\_MESH} / \texttt{ERR\_TIMEOUT}). Each retry is a fresh API call --- not a chat continuation --- whose user message contains the original task, the previous attempt's full Python code, and the truncated stderr/traceback (head-$70\%$ + tail-$30\%$ within a $3$\, K-character cap). Each backbone retains its peak thinking level across the multi-turn pass: Gemini and Gemma at \texttt{high}, Claude at \texttt{low} (each model's empirical peak), GPT-5.4-mini / GPT-5.4 at \texttt{medium}, and GPT-5.5 at \texttt{high}. Per-cell results are in Appendix~\ref{app:multi-turn-debug}. The complementary \emph{coding-agent harness} regime (Finding~5 in the main text) wraps each backbone in its provider-native CLI agent --- Claude~Code for Sonnet/Opus, OpenAI Codex CLI for the GPT family, and Google's \texttt{gemini-cli} for the Gemini family --- and gives the harness a per-instance wall-clock budget of $600$--$900$\,s to autonomously write, run, debug, and iterate on a Blender script in a sandbox directory. Gemma is omitted from the agentic-harness regime because it has no first-party CLI agent.

\subsection{Thinking-Level Ablation: 3D-Shape and DINOv3 Metrics}
\label{app:thinking-ablation}

Figure~\ref{fig:abl-combined}(a,b) in the main paper sweeps the thinking budget under SigLIP-2 view-paired (penalized). Here we complement that view with four additional metrics computed on the same set of cells, in the same per-family palette: Uni3D 3D--3D paired cosine and Uni3D cross-modal cosine on both tracks, and DINOv3 view-paired cosine on the image-to-3D track (text-to-3D has no rendered reference views, so it carries the Uni3D text--3D variant instead). Every cell is the penalized mean across the $212$ instances, with failed renders contributing $0$; Gemini- and Gemma-family rows are 3-seed mean $\pm$ std at seeds $\{0, 1, 2\}$, and Claude / GPT rows are single-seed under default API sampling.

\begin{figure}[htbp]
  \centering
  \includegraphics[width=0.99\linewidth]{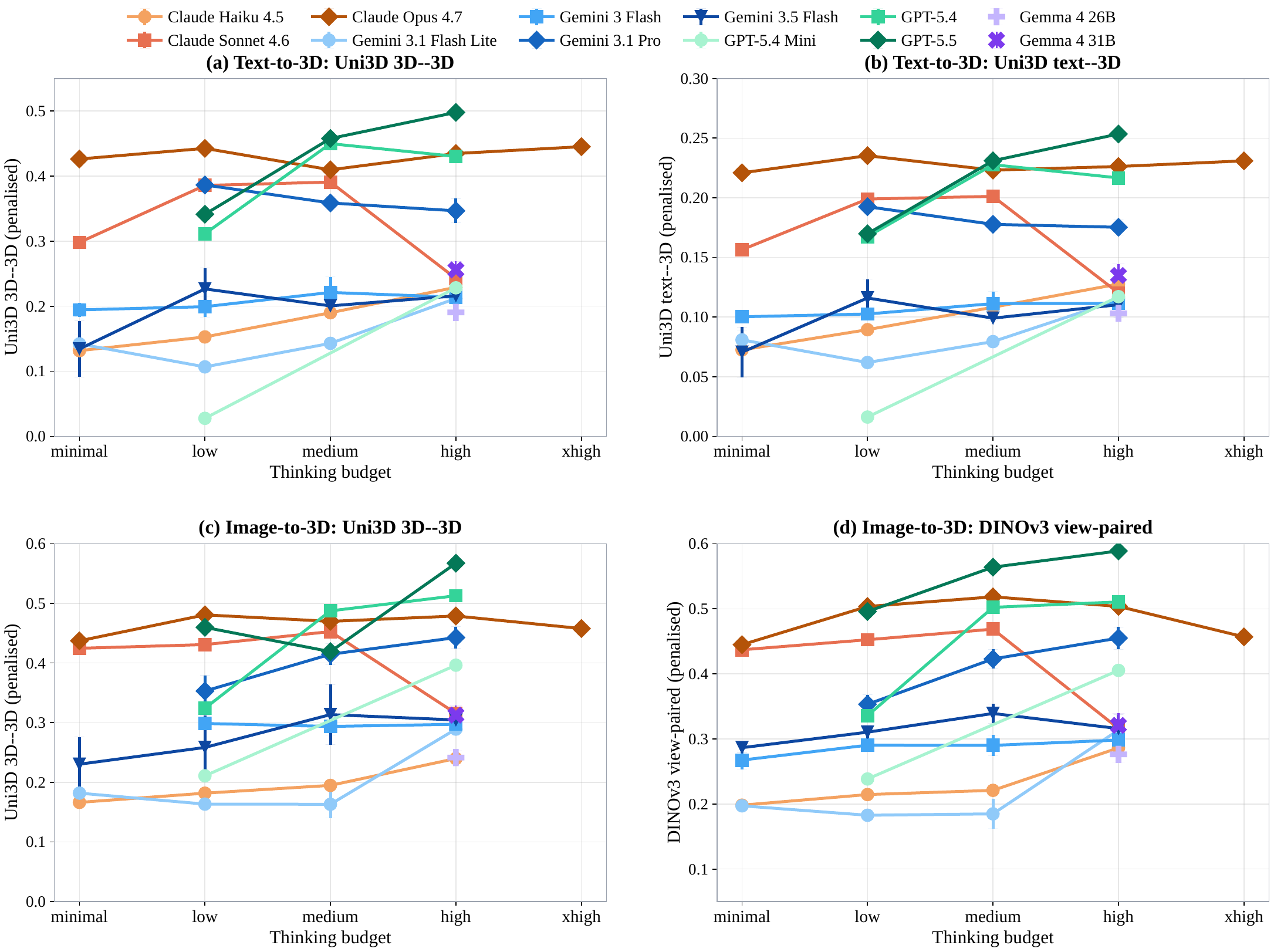}
  \caption{\textbf{Thinking-level ablation, complementary metrics.} Top row: text-to-3D, with (a) Uni3D 3D--3D paired cosine and (b) Uni3D text--3D cross-modal cosine (the input prompt vs.\ the generated point cloud). Bottom row: image-to-3D, with (c) Uni3D 3D--3D paired cosine and (d) DINOv3 view-paired (\texttt{facebook/dinov3-vitl16-pretrain-lvd1689m}) cosine. All values are penalized means; error bars are $1\sigma$ across $3$ Gemini/Gemma seeds.}
  \label{fig:abl-thinking-appendix}
\end{figure}

The qualitative picture matches the SigLIP-2 figure in the main paper. Lightweight reasoners (Gemini~3.1~Flash~Lite, Claude~Haiku~4.5, GPT-5.4-mini) gain substantially across \texttt{minimal}$\to$\texttt{high} on every metric and panel; frontier reasoners (Claude~Opus~4.7, Gemini~3.1~Pro, GPT-5.5) plateau early, with Opus already at the per-track ceiling at \texttt{minimal}; and Pro flips between tracks (text decreases mildly, image increases sharply) consistent with the headline single-turn ablation finding. The Uni3D 3D--3D panels (a, c) show the same monotone-with-step lift on Flash~Lite that SigLIP-2 captures, confirming that the Flash~Lite step is a genuine geometric improvement and not just a render-side similarity artifact; the DINOv3 panel (d) tracks SigLIP-2 closely, as expected from the strong DINOv3$\leftrightarrow$SigLIP-2 correlation in the main paper's perception--Elo analysis.

\subsection{Multi-View Image Budget Ablation}
\label{app:images-amount-ablation}

\textbf{Setup.} The image-to-3D track in the main results sends all four reference views (azimuths $45^{\circ}/135^{\circ}/225^{\circ}/315^{\circ}$) to the model. We ask whether this multi-view budget actually helps, or whether a single canonical front view already recovers the geometry. We sweep the number of input views $N \in \{1, 2, 3, 4\}$ in canonical order (the $45^{\circ}$ view first, then progressively adding $135^{\circ}$, $225^{\circ}$, $315^{\circ}$) on the six open-budget image-to-3D backbones (Gemini~3~Flash, Gemini~3.1~Flash~Lite, Gemini~3.1~Pro, Gemini~3.5~Flash, Gemma~4~26B, Gemma~4~31B) at \texttt{thinking}=\texttt{high}, with $3$ seeds per cell. All cells share the same image-to-3D system prompt (Appendix~\ref{app:sysprompt-image}), \texttt{temperature}=$0.7$, and the same render and metric pipelines as the main results (Section~\ref{sec:experiments}); the $N{=}4$ rows are the three \texttt{thinking}=\texttt{high} seeds reused from the thinking-level ablation, so the four cells per backbone differ only in the number of input views.

\begin{table}[htbp]
\centering
\caption{Image-to-3D multi-view budget ablation. We vary the
number of canonical reference views $N$ sent to the model, in
order $\{005, 015, 025, 035\}$, at
\texttt{thinking}=\texttt{high}. Each cell is mean$\pm$std
across seeds $\{0,1,2\}$. SigLIP-2 / DINOv3 columns are
view-paired image--image cosine between generated and reference
renders; Uni3D 3D--3D is paired point-cloud cosine between the
generated and reference GLBs, and Uni3D image--3D is the
cross-modal cosine between the canonical reference image and the
generated point cloud. All four similarity columns are penalized
so that failed runs contribute $0$.
\textbf{Bold} marks the best $N$ within each model on each
metric.}
\label{tab:images-amount}
\small
\setlength{\tabcolsep}{4pt}
\begin{tabular}{l c ccccc}
\toprule
Model & $N$
 & Exec.\,$\uparrow$
 & SigLIP-2\,$\uparrow$
 & DINOv3\,$\uparrow$
 & Uni3D 3D--3D\,$\uparrow$
 & Uni3D image--3D\,$\uparrow$ \\
\midrule
Gemini 3 Flash         & 1 & \msd{0.535}{0.010} & \msd{0.431}{0.007} & \msd{0.287}{0.001} & \msd{0.056}{0.001} & \msd{0.037}{0.001} \\
Gemini 3 Flash         & 2 & \msd{0.574}{0.012} & \msd{0.465}{0.011} & \msd{0.312}{0.008} & \textbf{\msd{0.060}{0.004}} & \textbf{\msd{0.042}{0.002}} \\
Gemini 3 Flash         & 3 & \textbf{\msd{0.586}{0.018}} & \textbf{\msd{0.475}{0.019}} & \textbf{\msd{0.320}{0.016}} & \msd{0.056}{0.004} & \msd{0.041}{0.002} \\
Gemini 3 Flash         & 4 & \msd{0.539}{0.020} & \msd{0.436}{0.015} & \msd{0.298}{0.008} & \msd{0.055}{0.002} & \msd{0.039}{0.002} \\
\addlinespace
Gemini 3.5 Flash       & 1 & \msd{0.476}{0.021} & \msd{0.420}{0.013} & \msd{0.293}{0.003} & \msd{0.184}{0.119} & \msd{0.103}{0.062} \\
Gemini 3.5 Flash       & 2 & \msd{0.513}{0.005} & \msd{0.440}{0.007} & \msd{0.313}{0.006} & \msd{0.198}{0.135} & \msd{0.107}{0.067} \\
Gemini 3.5 Flash       & 3 & \msd{0.483}{0.030} & \msd{0.425}{0.017} & \msd{0.308}{0.017} & \msd{0.198}{0.134} & \msd{0.106}{0.067} \\
Gemini 3.5 Flash       & 4 & \msd{0.514}{0.026} & \msd{0.445}{0.035} & \msd{0.316}{0.026} & \textbf{\msd{0.305}{0.023}} & \textbf{\msd{0.158}{0.011}} \\
\addlinespace
Gemini 3.1 Flash Lite  & 1 & \msd{0.624}{0.027} & \msd{0.484}{0.018} & \msd{0.306}{0.011} & \msd{0.058}{0.002} & \msd{0.045}{0.002} \\
Gemini 3.1 Flash Lite  & 2 & \msd{0.612}{0.018} & \msd{0.474}{0.012} & \msd{0.303}{0.006} & \msd{0.059}{0.001} & \msd{0.045}{0.000} \\
Gemini 3.1 Flash Lite  & 3 & \msd{0.624}{0.024} & \msd{0.487}{0.022} & \msd{0.309}{0.016} & \msd{0.058}{0.003} & \msd{0.045}{0.003} \\
Gemini 3.1 Flash Lite  & 4 & \textbf{\msd{0.627}{0.016}} & \textbf{\msd{0.493}{0.009}} & \textbf{\msd{0.314}{0.001}} & \textbf{\msd{0.062}{0.002}} & \textbf{\msd{0.048}{0.001}} \\
\addlinespace
Gemini 3.1 Pro         & 1 & \msd{0.731}{0.021} & \msd{0.601}{0.022} & \msd{0.431}{0.017} & \msd{0.070}{0.001} & \msd{0.050}{0.001} \\
Gemini 3.1 Pro         & 2 & \msd{0.753}{0.036} & \msd{0.622}{0.033} & \textbf{\msd{0.455}{0.026}} & \msd{0.073}{0.002} & \msd{0.052}{0.002} \\
Gemini 3.1 Pro         & 3 & \msd{0.744}{0.012} & \msd{0.610}{0.012} & \msd{0.440}{0.014} & \textbf{\msd{0.074}{0.001}} & \msd{0.052}{0.002} \\
Gemini 3.1 Pro         & 4 & \textbf{\msd{0.758}{0.031}} & \textbf{\msd{0.632}{0.022}} & \msd{0.455}{0.021} & \msd{0.074}{0.000} & \textbf{\msd{0.053}{0.001}} \\
\addlinespace
Gemma 4 26B            & 1 & \textbf{\msd{0.582}{0.021}} & \textbf{\msd{0.456}{0.011}} & \textbf{\msd{0.282}{0.007}} & \textbf{\msd{0.054}{0.002}} & \textbf{\msd{0.041}{0.002}} \\
Gemma 4 26B            & 2 & \msd{0.561}{0.077} & \msd{0.437}{0.061} & \msd{0.266}{0.040} & \msd{0.052}{0.005} & \msd{0.040}{0.004} \\
Gemma 4 26B            & 3 & \msd{0.550}{0.035} & \msd{0.426}{0.026} & \msd{0.263}{0.014} & \msd{0.050}{0.006} & \msd{0.038}{0.003} \\
Gemma 4 26B            & 4 & \msd{0.574}{0.031} & \msd{0.449}{0.022} & \msd{0.276}{0.014} & \msd{0.053}{0.002} & \msd{0.040}{0.002} \\
\addlinespace
Gemma 4 31B            & 1 & \msd{0.635}{0.031} & \msd{0.501}{0.028} & \msd{0.325}{0.026} & \msd{0.062}{0.002} & \msd{0.046}{0.001} \\
Gemma 4 31B            & 2 & \textbf{\msd{0.662}{0.036}} & \textbf{\msd{0.524}{0.030}} & \textbf{\msd{0.341}{0.019}} & \textbf{\msd{0.066}{0.002}} & \textbf{\msd{0.049}{0.002}} \\
Gemma 4 31B            & 3 & \msd{0.637}{0.017} & \msd{0.504}{0.013} & \msd{0.332}{0.009} & \msd{0.062}{0.003} & \msd{0.046}{0.001} \\
Gemma 4 31B            & 4 & \msd{0.605}{0.039} & \msd{0.485}{0.033} & \msd{0.321}{0.023} & \msd{0.061}{0.004} & \msd{0.045}{0.002} \\
\bottomrule
\end{tabular}
\end{table}

\textbf{Findings.} The six backbones show a clean capacity-ordered pattern.
Gemini~3.1~Pro is the only one that does not regress at $N{=}4$: it peaks at $N{=}4$ ($0.758\!\pm\!0.031$ Exec.) with $N{=}2$ a statistically tied runner-up, so for Pro every $N \in \{2,3,4\}$ is essentially indistinguishable and all three beat $N{=}1$ substantially.
As capacity drops, the optimum shifts left: Gemini~3~Flash peaks at $N{=}3$, Gemma~4~31B peaks at $N{=}2$, and Gemma~4~26B peaks at $N{=}1$ and is flat-to-degrading thereafter; only Gemini~3.1~Flash~Lite is view-budget-insensitive within seed noise.
We read this as: smaller backbones extract less marginal information from each additional view but incur the same context cost, so they saturate or get distracted earlier in the view budget, whereas only the largest backbone can absorb four reference views without trading off response quality elsewhere.

% =========================================================================
% [Qualitative comparison was moved into the main text as
%  Section~\ref{sec:qualitative} in sections/experiments.tex.]
\FloatBarrier
% =========================================================================
\section{Evaluation metric implementation}
\label{app:metrics}

Section~\ref{sec:protocol} defines the metric suite abstractly. This appendix records the released implementation. We reuse the notation of Section~\ref{sec:task}: a policy $\pi$ produces a script $f_\pi = \pi(c)$, which the deterministic Blender~5.0 operator $\mathcal{E}$ compiles into a mesh $M_\pi = \mathcal{E}(f_\pi)$. Our render driver runs $f_\pi$ in a fresh Blender~5.0 subprocess (wall-clock budget $240$~s) and renders $M_\pi$ from the four canonical views $V = \{45^\circ, 135^\circ, 225^\circ, 315^\circ\}$ matching frames $5/15/25/35$ of the reference Infinigen turntable. We write $r_v(M_\pi)$ for the render of $M_\pi$ at view $v$, $g_v$ for the corresponding reference Infinigen render, and use the per-instance subscript $i$ over $N=212$ test instances.

\subsection{Executability}
\label{app:metrics-exec}

The executability indicator for instance $i$ is
\begin{equation}
\mathrm{Exec}_i \;=\;
\mathbb{I}\!\bigl[\,
  \mathcal{E}(f_{\pi,i}) \neq \emptyset \;\wedge\;
  |\mathrm{Mesh}(\mathcal{E}(f_{\pi,i}))| \geq 1
\,\bigr],
\label{eq:exec}
\end{equation}
where $\mathrm{Mesh}(\cdot)$ counts mesh objects in the post-execution scene and the $240$~s timeout maps to $\mathcal{E}(f_\pi)\!=\!\emptyset$. The aggregate rate is $\overline{\mathrm{Exec}} = N^{-1}\!\sum_i \mathrm{Exec}_i$. Each failure is bucketed into one mutually exclusive stage --- \texttt{ERR\_EXEC} (Python exception), \texttt{ERR\_NO\_MESH} (no mesh in the resulting scene), \texttt{ERR\_RENDER} (one of the four views failed to render), or \texttt{ERR\_TIMEOUT} ($240$~s budget exceeded). The metric script also reports recurring exception fingerprints (a hash of the error type + a truncated traceback) to diagnose systematic API mismatches, such as removed Blender 4.x calls.

\subsection{Image-Grounded Similarity (Image-to-3D Track)}
\label{app:metrics-image-image}

For e,ach view we pair $r_v(M_{\pi,i})$ with $g_v$ at the same camera and compute cosine similarity under image encoder $\psi$:
\begin{equation}
\sigma^{\psi}_{i,v} \;=\; \cos\bigl(\psi(r_v(M_{\pi,i})),\;\psi(g_v)\bigr),
\qquad
\sigma^{\psi}_{i} \;=\; \frac{1}{|V|}\sum_{v \in V}\sigma^{\psi}_{i,v},
\label{eq:image-image}
\end{equation}
instantiated with $\psi_{\text{SigLIP-2}}$~\citep{tschannen2025siglip2} (\texttt{google/\allowbreak{}siglip2-\allowbreak{}so400m-\allowbreak{}patch16-\allowbreak{}naflex}, image branch --- semantic correspondence) and $\psi_{\text{DINOv3}}$~\citep{simeoni2025dinov3} (\texttt{facebook/\allowbreak{}dinov3-\allowbreak{}vitl16-\allowbreak{}pretrain-\allowbreak{}lvd1689m} ViT-L/16 --- shape and structural correspondence; less sensitive to surface appearance, which is desirable here because generated meshes are rendered untextured against a neutral background while the reference renders are full-color). Reporting both columns separately exposes the shape-vs-semantic trade-off (cf.\ Gemini~3~Flash vs Gemini~3.1~Flash~Lite in Table~\ref{tab:t1}). We omit the $\max_v$ aggregation: under view-paired comparison, it reduces to ``the easiest viewpoint to match'', which a model can satisfy without recovering the overall geometry.

\subsection{Text-Render Similarity (Text-to-3D Track)}
\label{app:metrics-siglip}

The text-to-3D track replaces $\psi(g_v)$ with the SigLIP-2 \emph{text} embedding of the prompt $c_i$, with the same image branch $\phi_{\text{img}} = \psi_{\text{SigLIP-2}}$ and matched text branch $\phi_{\text{txt}}$:
\begin{equation}
s^{\text{mean}}_i \;=\; \frac{1}{|V|}\sum_{v \in V}\cos\bigl(\phi_{\text{img}}(r_v(M_{\pi,i})),\; \phi_{\text{txt}}(c_i)\bigr),
\quad
s^{\text{max}}_i \;=\; \max_{v \in V}\;\cos\bigl(\phi_{\text{img}}(r_v(M_{\pi,i})),\; \phi_{\text{txt}}(c_i)\bigr).
\label{eq:siglip}
\end{equation}
We additionally report a \emph{GT baseline} computed by applying the same metric to the four reference Infinigen renders for the same prompts; this is a soft ceiling that no model can be expected to substantially exceed without exploiting prompt-level shortcuts.

\subsection{3D-Shape Similarity}
\label{app:metrics-3d}

We additionally score each executable instance directly on the exported GLB. Let $P_{\pi,i} = \mathrm{sample}_K(M_{\pi,i})$ and $P^\star_i = \mathrm{sample}_K(M^\star_i)$ denote $K=8192$ surface-sampled points (uniform area sampling), each independently centered at the centroid and rescaled to the unit bounding sphere.

\textbf{Chamfer Distance.} We report symmetric squared Chamfer~\citep{fan2017psgn} with $4$-yaw alignment to absorb canonical-orientation mismatch:
\begin{equation}
\mathrm{CD}_i \;=\;
\min_{\theta\,\in\,\{0^\circ,90^\circ,180^\circ,270^\circ\}}\!
\Bigl[\,
\frac{1}{K}\!\!\sum_{p \in P^\star_i}\!\min_{q\,\in\,R_\theta P_{\pi,i}}\!\|p-q\|_2^2
\;+\;
\frac{1}{K}\!\!\sum_{q\,\in\,R_\theta P_{\pi,i}}\!\min_{p \in P^\star_i}\!\|p-q\|_2^2
\,\Bigr],
\label{eq:chamfer}
\end{equation}
where $R_\theta$ rotates around the world $z$-axis and nearest-neighbor queries use \texttt{cKDTree}.

\textbf{Uni3D 3D--3D and cross-modal cosine.} Following Uni3D~\citep{zhou2024uni3d}, we encode the (xyz, rgb) point cloud with the Uni3D-Giant point encoder $\eta_{\text{pc}}$ (\texttt{BAAI/Uni3D: modelzoo/uni3d-g}; EVA-Giant backbone, $K=8192$ points $+$ $3$-channel RGB) into the $1024$-dim CLIP-aligned latent shared with the EVA02-E-14-plus text and image branches $\eta_{\text{txt}}, \eta_{\text{img}}$ (\texttt{laion2b\_s9b\_b144k}). The two reported similarities are
\begin{equation}
u^{\text{3D}}_i \;=\; \cos\bigl(\eta_{\text{pc}}(P_{\pi,i}),\;\eta_{\text{pc}}(P^\star_i)\bigr),
\qquad
u^{\text{xm}}_i \;=\;
\begin{cases}
\cos\bigl(\eta_{\text{pc}}(P_{\pi,i}),\;\eta_{\text{txt}}(c_i)\bigr) & \text{text-to-3D},\\[2pt]
\cos\bigl(\eta_{\text{pc}}(P_{\pi,i}),\;\eta_{\text{img}}(g_{45^\circ,i})\bigr) & \text{image-to-3D},
\end{cases}
\label{eq:uni3d}
\end{equation}
i.e.\ the cross-modal column compares the generated shape against either the prompt or the canonical $45^\circ$ reference view, depending on the task.

\subsection{Conditional vs.\ Penalized Aggregation}
\label{app:metrics-policy}

For any per-instance quality metric $\mathcal{D}_i \in \{s^{\text{mean}}_i,\, s^{\text{max}}_i,\, \sigma^{\text{SigLIP-2}}_i,\, \sigma^{\text{DINOv3}}_i,\, \mathrm{CD}_i,\, u^{\text{3D}}_i,\, u^{\text{xm}}_i\}$ we report two cross-instance aggregations:
\begin{equation}
\overline{\mathcal{D}}^{\text{cond}} \;=\; \frac{\sum_{i=1}^{N}\mathrm{Exec}_i\cdot\mathcal{D}_i}{\sum_{i=1}^{N}\mathrm{Exec}_i},
\qquad
\overline{\mathcal{D}}^{\text{pen}} \;=\; \frac{1}{N}\sum_{i=1}^{N}\mathrm{Exec}_i\cdot \mathcal{D}_i,
\label{eq:cond-pen}
\end{equation}
with the convention that for the lower-is-better Chamfer Distance, $\overline{\mathrm{CD}}^{\text{pen}}$ assigns the run-relative penalty $1.5\cdot\!\max_{i:\mathrm{Exec}_i=1}\mathrm{CD}_i$ (rather than $0$) to failed instances, keeping the metric finite while preserving its lower-is-better semantics. The \emph{conditional} form isolates geometric quality from code reliability; the \emph{penalized} form contributes $0$ (or the Chamfer penalty) for failed instances, so a model is never rewarded for suppressing weak outputs. Headline numbers in Table~\ref{tab:t1} are penalized; ablations vary by what is more informative and state the choice per table.

\FloatBarrier
% =========================================================================
\section{Inference setup}
\label{app:inference}

Each task uses its own system prompt, reproduced verbatim below.
The text-to-3D system prompt (Appendix~\ref{app:sysprompt-text}) is paired at inference time with the per-instance \texttt{prompt\_description.txt} as a user message; the image-to-3D system prompt (Appendix~\ref{app:sysprompt-image}) is paired with the four reference renders.
Both prompts fix the same three things in turn: the strict output format (raw Blender~5.0 Python, no Markdown), the target environment (Blender~5.0 with a closed allow-list of libraries), and the behavioral code requirements (single object, no scenery, deterministic, no rendering, no file I/O).

\subsection{Text-to-3D System Prompt}
\label{app:sysprompt-text}

\begin{spromptbox}[Text-to-3D system prompt]{spromptBlue}{spromptBlueBg}
\small
You are a procedural 3D modeling expert. Given a
natural-language description of a 3D object, you produce a
single self-contained Python script that constructs the
described object in Blender~5.0.

\medskip
\noindent\textbf{\# Output format (HARD constraint -- read carefully)}

\smallskip
Your entire response will be saved verbatim into a \texttt{.py}
file and executed by Blender~5.0. ANY non-Python content
anywhere in the response will break that file. Treat this as a
strict machine-to-machine contract, not a chat reply.

\smallskip
Your response MUST consist solely of Python source code.
Specifically:
\begin{itemize}[leftmargin=1.2em,itemsep=1pt,topsep=2pt]
  \item Do NOT emit Markdown code fences anywhere -- no opening
    \texttt{```python}, no opening \texttt{```py}, no opening
    \texttt{```}, no closing \texttt{```} at the end. Not even
    one. Not even on a single line by itself.
  \item Do NOT prepend any prose (``Here is the code'', ``I
    will create'', ``This script generates'', ``Sure'', etc.).
  \item Do NOT append any prose (``Hope this helps'', ``Let me
    know if'', ``Note that\dots'', etc.).
  \item Do NOT include HTML/XML tags, bullet points, headings,
    or any formatting other than plain Python.
  \item The very first character of your response must be the
    first character of a valid Python source (typically
    \texttt{i} of \texttt{import bpy}, or \texttt{\#} of a
    top-level comment, or \texttt{"} of a module docstring).
  \item The very last character of your response must be the
    last character of the Python script (a closing parenthesis,
    a bare newline at the end of the file, etc.) -- never a backtick.
\end{itemize}

\smallskip
\textbf{Self-check before answering:} if I save your response
to \texttt{out.py} and run\\
\texttt{python -c "import ast; ast.parse(open('out.py').read())"}\\
It must succeed without modification.

\medskip
\noindent\textbf{\# Target environment}

\smallskip
\begin{itemize}[leftmargin=1.2em,itemsep=1pt,topsep=2pt]
  \item Blender version: \textbf{5.0}. Use the Blender 5.0
    Python API (\texttt{bpy}, \texttt{bmesh},
    \texttt{mathutils}).
  \item The script will be executed via
    \texttt{blender --background --python <file>} or pasted
    into the Blender Text Editor.
  \item Allowed libraries (the ONLY libraries you may import):
    \begin{itemize}[leftmargin=1.2em,itemsep=0pt,topsep=1pt]
      \item Blender-bundled: \texttt{bpy}, \texttt{bmesh},
        \texttt{mathutils}.
      \item Python stdlib: \texttt{math}, \texttt{random},
        \texttt{itertools}, \texttt{collections},
        \texttt{functools}, \texttt{dataclasses},
        \texttt{enum}, \texttt{typing}.
      \item Numerical: \texttt{numpy}, \texttt{scipy}.
    \end{itemize}
  \item Do NOT import anything that requires network access,
    GUI interaction, or external file reads (no \texttt{os},
    \texttt{sys.path} hacks, \texttt{requests}, \texttt{PIL},
    \texttt{cv2}, \dots).
\end{itemize}

\medskip
\noindent\textbf{\# Code requirements}

\smallskip
\begin{itemize}[leftmargin=1.2em,itemsep=1pt,topsep=2pt]
  \item Produce ONE final 3D object (or coherent assembly)
    that corresponds to the description.
  \item Generate ONLY the described object -- NO ground plane,
    NO backdrop, NO skybox, NO environmental props, NO
    decorative context. No grass under the chair, no pedestal
    beneath the figurine, no ``studio floor'' plane. After
    your script finishes, the scene must contain only your
    geometry, sitting at the origin.
  \item Push for as much geometric detail as the description.
     If the prompt mentions ribs, slats, vents,
    handles, fins, scales, leaves, rivets, pleating,
    segmentation, or ornament, model them as real geometry
    rather than as flat surfaces with a label. Use subdivision
    surface, bevel, array, mirror, and screw modifiers when
    they are the right tool, and drop into \texttt{bmesh} for
    finer features. Aim for high fidelity but keep meshes
    within a few hundred thousand vertices -- the renderer has
    a 240~s budget per script.
  \item Prefer procedural construction: parametric loops,
    \texttt{bmesh} operators, modifiers, and array/mirror
    operations. Avoid hard-coded long vertex lists.
  \item At the start of the script, clear the default scene
    (delete the default cube, camera, and light if present)
    so that the output scene contains only your generated
    geometry.
  \item Leave geometry untextured.
  \item No need to save the \texttt{.blend} file. Do NOT
    trigger a render. Do NOT call \texttt{sys.exit} or
    \texttt{bpy.ops.wm.quit\_blender}.
  \item If the description is ambiguous on dimensions,
    proportions, or stylistic details, choose reasonable
    defaults silently and proceed. Never ask clarifying
    questions.
  \item The script must terminate normally; final geometry
    must exist in \texttt{bpy.data.objects} when execution
    completes.
\end{itemize}
\end{spromptbox}

\subsection{Image-to-3D system prompt}
\label{app:sysprompt-image}

\begin{spromptbox}[Image-to-3D system prompt]{spromptBlue}{spromptBlueBg}
\small
You are a procedural 3D modeling expert. Given one or more
reference images of a 3D object, you produce a single
self-contained Python script that reconstructs the depicted
object in Blender~5.0.

\medskip
\noindent\textbf{\# Input format}

\smallskip
You will receive one or more reference images of a single
target object as part of the user message. The images may be:
\begin{itemize}[leftmargin=1.2em,itemsep=1pt,topsep=2pt]
  \item A single view (front, side, 3/4, etc.) -- infer unseen
    sides by symmetry and category priors.
  \item Multiple views of the SAME object (e.g.\ front + side
    + back, or turntable frames). Treat them as multi-view
    evidence of one object, not as separate objects.
    Cross-reference views to resolve depth, proportions, and
    occluded structure.
  \item A mix of full-object shots and close-up detail crops.
    Use the close-ups to refine local geometry (handles,
    vents, ornament) of the same object.
\end{itemize}

\smallskip
If the images appear to depict different objects, model the
most prominent / first-shown object and silently ignore the
rest.

\smallskip
Read the images carefully before writing code. Identify the
object's category, overall proportions
(length:width:height), part decomposition, symmetries
(bilateral, radial, none), repeating structures (slats, ribs,
teeth, scales, leaves), and any distinctive ornament. Your
script should reproduce these as real geometry.

\medskip
\noindent\textbf{\# Output format (HARD constraint -- read carefully)}

\smallskip
Your entire response will be saved verbatim into a \texttt{.py}
file and executed by Blender~5.0. ANY non-Python content
anywhere in the response will break that file. Treat this as a
strict machine-to-machine contract, not a chat reply.

\smallskip
Your response MUST consist entirely of Python source code.
Specifically:
\begin{itemize}[leftmargin=1.2em,itemsep=1pt,topsep=2pt]
  \item Do NOT emit Markdown code fences anywhere -- no opening
    \texttt{```python}, no opening \texttt{```py}, no opening
    \texttt{```}, no closing \texttt{```} at the end. Not even
    one. Not even on a single line by itself.
  \item Do NOT prepend any prose (``Here is the code'', ``I
    will create'', ``This script generates'', ``Sure'',
    ``Looking at the image'', etc.).
  \item Do NOT append any prose (``Hope this helps'', ``Let me
    know if'', ``Note that\dots'', etc.).
  \item Do NOT include HTML/XML tags, bullet points, headings,
    or any formatting other than plain Python.
  \item Do NOT describe what you see in the image as prose; if
    you want to record observations, put them in Python
    comments inside the script.
  \item The very first character of your response must be the
    first character of valid Python source (typically
    \texttt{i} of \texttt{import bpy}, or \texttt{\#} of a
    top-level comment, or \texttt{"} of a module docstring).
  \item The very last character of your response must be the
    last character of the Python script (a closing parenthesis,
    a bare newline at end of file, etc.) -- never a backtick.
\end{itemize}

\smallskip
\textbf{Self-check before answering:} if I save your response
to \texttt{out.py} and run\\
\texttt{python -c "import ast; ast.parse(open('out.py').read())"}\\
it must succeed without modification.

\medskip
\noindent\textbf{\# Target environment}

\smallskip
\begin{itemize}[leftmargin=1.2em,itemsep=1pt,topsep=2pt]
  \item Blender version: \textbf{5.0}. Use Blender 5.0
    Python API (\texttt{bpy}, \texttt{bmesh},
    \texttt{mathutils}).
  \item The script will be executed via
    \texttt{blender --background --python <file>} or pasted
    into the Blender Text Editor.
  \item Allowed libraries (the ONLY libraries you may import):
    \begin{itemize}[leftmargin=1.2em,itemsep=0pt,topsep=1pt]
      \item Blender-bundled: \texttt{bpy}, \texttt{bmesh},
        \texttt{mathutils}.
      \item Python stdlib: \texttt{math}, \texttt{random},
        \texttt{itertools}, \texttt{collections},
        \texttt{functools}, \texttt{dataclasses},
        \texttt{enum}, \texttt{typing}.
      \item Numerical: \texttt{numpy}, \texttt{scipy}.
    \end{itemize}
  \item Do NOT import anything that requires network access,
    GUI interaction, or external file reads (no \texttt{os},
    \texttt{sys.path} hacks, \texttt{requests}, \texttt{PIL},
    \texttt{cv2}, \dots). In particular, the reference images
    are NOT available to the script at runtime -- do not
    attempt to load, open, or read them from disk; encode
    everything you inferred from the images directly into the
    geometry your script builds.
\end{itemize}

\medskip
\noindent\textbf{\# Code requirements}

\smallskip
\begin{itemize}[leftmargin=1.2em,itemsep=1pt,topsep=2pt]
  \item Produce ONE final 3D object (or coherent assembly)
    that corresponds to the object shown in the reference
    images.
  \item Generate ONLY the depicted object -- NO ground plane,
    NO backdrop, NO skybox, NO environmental props, NO
    decorative context, even if the reference images show a
    background, a floor, a stand, a hand holding the object,
    or other surrounding scenery. Model only the object
    itself, sitting at the origin.
  \item Match the reference as faithfully as the geometry
    allows: overall silhouette, part proportions, count and
    placement of repeating elements (e.g.\ number of legs,
    number of petals, number of slats), and characteristic
    curvature. When the images give clear quantitative cues
    (e.g.\ ``8 spokes'', ``3 drawers''), reproduce those
    counts exactly rather than approximating.
  \item Push for as much geometric detail as the images
    warrant. If the reference shows ribs, slats, vents,
    handles, fins, scales, leaves, rivets, pleating,
    segmentation, or ornament, model them as real geometry
    rather than as flat surfaces with a label. Use subdivision
    surface, bevel, array, mirror, and screw modifiers when
    they are the right tool, and drop into \texttt{bmesh} for
    finer features. Aim for high fidelity but keep meshes
    within a few hundred thousand vertices -- the renderer has
    a 240~s budget per script.
  \item Prefer procedural construction: parametric loops,
    \texttt{bmesh} operators, modifiers, and array/mirror
    operations. Avoid hard-coded long vertex lists.
  \item Exploit symmetry when the reference supports it
    (mirror modifier for bilateral objects, array/spin for
    radial ones); this both saves vertices and makes the
    result cleaner.
  \item At the start of the script, clear the default scene
    (delete the default cube, camera, and light if present)
    so that the output scene contains only your generated
    geometry.
  \item Leave geometry untextured. Do NOT try to reproduce the
    colors, materials, or surface texture seen in the
    reference -- geometry only. Ignore lighting, shadows, and
    photographic artifacts in the images; they are not part
    of the object.
  \item No need to save the \texttt{.blend} file. Do NOT
    trigger a render. Do NOT call \texttt{sys.exit} or
    \texttt{bpy.ops.wm.quit\_blender}.
  \item If the images are ambiguous about back-side details,
    internal structure, or absolute scale, choose reasonable
    defaults silently (consistent with the visible views and
    the object category) and proceed. Never ask clarifying
    questions.
  \item The script must terminate normally; final geometry
    must exist in \texttt{bpy.data.objects} when execution
    completes.
\end{itemize}
\end{spromptbox}

% -------------------------------------------------------------------------
\subsection{Multi-Turn Error-Feedback User Template}
\label{app:prompt-multiturn}

The multi-turn loop of Section~\ref{app:multi-turn-debug} reuses the original task's system prompt verbatim and replaces the user message with the template below. Each retry is stateless --- the previous attempt's code and the truncated Blender stderr (head $70\%$ + tail $30\%$ within $3$\,K characters) are pasted into the user message rather than passed as conversation history.

\begin{spromptbox}[Multi-turn user template]{spromptRed}{spromptRedBg}
\small
Your previous Blender~5.0 Python script for the task below FAILED to produce a valid render. Read the error carefully and output a corrected, complete script.

\medskip
\noindent\textbf{\# Original task}\\
\texttt{\{original\_task\_block\}}

\medskip
\noindent\textbf{\# Your previous code}\\
\texttt{\{prev\_code\}}

\medskip
\noindent\textbf{\# Execution result}
\begin{itemize}[leftmargin=1.5em,itemsep=0pt,topsep=2pt]
  \item status: \texttt{\{status\}}
  \item meshes produced: \texttt{\{n\_meshes\}}
  \item views rendered: \texttt{\{n\_views\_rendered\}}/4
  \item attempt: \texttt{\{attempt\_num\}} of \texttt{\{max\_attempts\}}
\end{itemize}

\medskip
\noindent\textbf{\# Error / diagnostic output (from Blender stderr/runtime)}\\
\texttt{\{error\_text\}}

\medskip
\noindent\textbf{\# Your task}\\
Fix the bug and output the COMPLETE corrected Python script. Do NOT output a diff, patch, or explanation. The same hard format constraints from the system prompt still apply: pure Python source only, no markdown fences, no prose before or after the code.

\medskip
Common failure modes to consider, depending on the error: \texttt{ERR\_EXEC} (Blender~5.0 API mismatch, missing import, wrong arg name); \texttt{ERR\_NO\_MESH} (only curves/empties/lights remain --- ensure at least one MESH object); \texttt{ERR\_RENDER} (NaN coordinates or infinite-extent geometry); \texttt{ERR\_TIMEOUT} (reduce subdivisions, runaway loops, heavy boolean ops). Do not just retry the same approach.
\end{spromptbox}

% -------------------------------------------------------------------------
\subsection{Visual Self-Critique Prompts}
\label{app:prompt-visualfb}

The visual self-critique loop of Section~\ref{app:visual-feedback} runs on \emph{baseline-OK} instances. The text-to-3D and image-to-3D variants share the same response-format contract (\texttt{NEEDS\_FIX:NO} or \texttt{NEEDS\_FIX:YES}\,/\,\texttt{<assessment>}\,/\,\texttt{<code>}) but differ in what the model is asked to compare against.

\begin{spromptbox}[Visual self-critique system prompt --- text-to-3D]{spromptTeal}{spromptTealBg}
\small
You are a 3D modeling expert reviewing your own previous work in Blender~5.0. You will be given a natural-language description, the Python script you previously wrote, up to $4$ rendered views, and an iteration counter.

\medskip
\noindent\textbf{\# Priority issues (in this order)}
\begin{enumerate}[leftmargin=1.6em,itemsep=0pt,topsep=2pt]
  \item Missing parts; \item Floating / disconnected pieces; \item Wrong proportions; \item Misalignment; \item Missing geometric detail; \item Wrong overall shape.
\end{enumerate}

\medskip
\noindent\textbf{\# Response format (HARD constraint)}\\
Either \texttt{NEEDS\_FIX:NO} + a one-sentence \texttt{<assessment>} (Pattern~A, no code), or \texttt{NEEDS\_FIX:YES} + a $\le$$100$-word bulleted \texttt{<assessment>} + a complete corrected Python script in \texttt{<code>...</code>} (Pattern~B). Hard rules for the corrected code: pure Python (no markdown fences inside the block); the first character is the first character of valid Python; complete self-contained script (not a diff); Blender~5.0 API only with the same closed allow-list of libraries as the system prompt; clears the default scene; one final MESH object at the origin; no rendering, no \texttt{.blend} save, no \texttt{sys.exit}; address the issues you listed.

\medskip
\noindent\textbf{\# What ``good enough'' means}\\
You have a limited iteration budget. If only minor proportions are off but the object is recognizable and complete, prefer \texttt{NEEDS\_FIX:NO}. If the render shows almost nothing, say \texttt{NEEDS\_FIX:YES} and rewrite from scratch.
\end{spromptbox}

\begin{spromptbox}[Visual self-critique system prompt --- image-to-3D (v1)]{spromptTeal}{spromptTealBg}
\small
The original task is \emph{image-to-3D reconstruction}: you were given $4$ reference images at $4$ turntable angles and asked to write a Blender~5.0 Python script that reproduces them. You will now be given the $4$ ORIGINAL reference images, your previous script, and $4$ RENDERED views from the same camera angles.

\medskip
\noindent\textbf{\# Comparison protocol.}\\
Compare your renders to the references \emph{view-by-view at the same angle} (frame $N$ of your render $\leftrightarrow$ frame $N$ of the reference). The reference images are the ground truth—they define what the object looks like.

\medskip
\noindent\textbf{\# Priority issues (in this order)}\\
Missing parts; wrong silhouette; wrong proportions; misalignment; floating pieces; missing geometric detail.

\medskip
\noindent\textbf{\# Response format.}\\
Same Pattern~A / Pattern~B contract as the text-to-3D critique prompt; the bulleted assessment in Pattern~B should name concrete differences between render and reference (``the references show $4$ legs but my render only has $2$'').

\medskip
\noindent\textbf{\# Conservatism bias.}\\
When in doubt, prefer \texttt{NEEDS\_FIX:NO}: each FIX risks breaking working code, and a ``good enough'' render that matches roughly is better than a rewritten render that doesn't render at all.
\end{spromptbox}

The user message for both variants supplies the $4$ renders (and, for image-to-3D, the $4$ references) plus an iteration counter \texttt{\{iter\_num\}/\{max\_iter\}} and a copy of the previous \texttt{\{prev\_code\}}.

% -------------------------------------------------------------------------
\subsection{Text-to-Image Generation Meta-Prompt}
\label{app:prompt-t2i}

For the text-to-image-to-3D pipeline of Section~\ref{app:text-to-image-to-3d}, each text description is first run through the meta-prompt below to produce a single studio-style reference photo that the image-to-3D code-generator then ingests under the standard image-to-3D system prompt.

\begin{spromptbox}[Text-to-image meta-prompt]{spromptGray}{spromptGrayBg}
\small
Generate ONE photographic reference image of the described 3D object, suitable as input for 3D reconstruction.

\medskip
\noindent\textbf{\# Required style}
\begin{itemize}[leftmargin=1.5em,itemsep=0pt,topsep=2pt]
  \item Single object, centered, occupying $\sim\!70$--$80\%$ of the frame.
  \item Plain neutral background (light gray, off-white, or soft gradient). No scene, no environment, no floor, no horizon, no wall.
  \item Soft, even, diffuse lighting from above-front; no harsh shadows, no dramatic side-lighting.
  \item Three-quarter front view ($\sim\!30$--$45^\circ$ around the vertical axis, $\sim\!15^\circ$ above eye level). The object should clearly read as a 3D form with depth visible.
  \item Neutral, true-to-life colors; no stylization, no painterly effects, no artistic filters.
  \item No text, labels, captions, logos, rulers, callouts; no people, animals, or hands; no exploded views, schematics, or multi-panel layouts.
  \item Photorealistic studio product-photography aesthetic.
\end{itemize}

\medskip
\noindent\textbf{\# Object to render}\\
\texttt{\{description\}}
\end{spromptbox}

% -------------------------------------------------------------------------
\subsection{Use of Agent Harnesses}
\label{app:agentic-harness}

The coding-agent regime (Finding~5 in the main text) wraps each backbone in its provider-native command-line agent and gives the harness the same task description as the single-shot baseline, plus a writable scratch directory. The system and user prompts are the harness's own defaults --- we do not override them --- so reproducibility reduces to (i)~the harness binary and version, (ii)~the writable working directory contents at $t=0$, and (iii)~the user instruction we pipe in.

\begin{itemize}[leftmargin=1.5em,itemsep=2pt,topsep=2pt]
  \item \emph{Anthropic Sonnet/Opus} run under \textbf{Claude Code} (\texttt{claude} CLI). At $t=0$ the working directory contains an empty \texttt{out.py} placeholder and a \texttt{run\_blender.sh} helper that calls Blender~5.0 headless and prints the stderr/stdout. The user instruction is the per-instance task description plus ``write \texttt{out.py} and iterate on it until \texttt{run\_blender.sh} succeeds and the script outputs a MESH object''.
  \item \emph{OpenAI GPT-5.x} run under \textbf{Codex CLI} (\texttt{codex} CLI). Same scratch directory contents and same user instructions; tool-call cap and reasoning effort follow the harness defaults for the underlying model (\texttt{medium} for GPT-5.4/GPT-5.4-mini, \texttt{high} for GPT-5.5).
  \item \emph{Google Gemini~3.x} run under \textbf{\texttt{gemini-cli}}. Same scratch directory and instruction; the harness drives Gemini~3 Flash, Flash Lite, and $3.1$~Pro through the same loop.
\end{itemize}

We allot a per-instance wall-clock budget of $600$--$900$\,s for the harness to autonomously edit the script, invoke Blender, parse stderr, and iterate; the loop terminates when the harness emits a successful render or when the budget expires. Final \texttt{out.py} files are then evaluated through the same render+metric pipeline as the single-shot baseline, so harness and baseline scores are directly comparable. Gemma~4 has no first-party CLI agent and is therefore omitted from this regime.

\FloatBarrier
% =========================================================================
\section{Iterative inference and multi-stage pipelines}
\label{app:multi-stage}

The primary table runs each model zero-shot, single-turn: one inference call per instance, no retry on Blender failures, no use of the model's own renders.
This appendix isolates three orthogonal extensions of that minimal pipeline that we think are worth probing on a procedural-3D code-generation benchmark and reports negative or task-asymmetric results that we believe are useful failure-mode evidence for future work.
The three extensions are: (1)~\emph{multi-turn error-feedback retry}, where the model is shown the Blender stderr from its own failed run and asked to fix it; (2)~\emph{visual self-critique}, where the model is shown its own rendered output and asked to either accept or rewrite the script; and (3)~\emph{text-to-image-to-3D pipelining}, where a frontier text-to-image model (Nano Banana Pro) generates a photorealistic reference image as an intermediate step.
We run the three extensions across an overlapping but not identical model set, chosen for affordability and to cover both ends of the capability spectrum.
The error-feedback retry (Section~\ref{app:multi-turn-debug}) is run on \textbf{eleven} models spanning all four model families: six Gemini-family models (Gemini~3~Flash, Gemini~3.1~Flash~Lite, Gemini~3.1~Pro, Gemini~3.5~Flash, Gemma~4~26B, Gemma~4~31B); the two strongest Anthropic models (Claude~Sonnet~4.6, Claude~Opus~4.7) at thinking=\texttt{low} (their empirical peak); and the three OpenAI models GPT-5.4-mini, GPT-5.4 (both at thinking=\texttt{medium}, their peak), and GPT-5.5 (at thinking=\texttt{high}, its peak).
The visual self-critique loop (Section~\ref{app:visual-feedback}) and the text-to-image-to-3D pipeline (Section~\ref{app:text-to-image-to-3d}) currently retain only the original four-model subset; the third extension additionally includes Gemini~3.1~Pro because its large reasoning budget changes the qualitative result.

% -------------------------------------------------------------------------
\subsection{Multi-Turn Error-Feedback Retry}
\label{app:multi-turn-debug}

\textbf{Setup.}
For each instance whose single-turn render fails (\texttt{ERR\_EXEC} / \texttt{ERR\_NO\_MESH} / \texttt{ERR\_TIMEOUT}), we run up to two additional \emph{stateless} retries (no chat history). Each retry's user message follows the template in Appendix~\ref{app:prompt-multiturn} and contains the original task, the previous full script, and the truncated stderr (head $70\%$ + tail $30\%$ within a $3$\,K-character cap). The loop stops on the first successful render or after the third attempt; baseline-OK instances are not touched.

\textbf{Findings.}
Table~\ref{tab:multi-turn-debug} reports executability lift, recovery decomposition, and --- crucially --- the change in \emph{penalized mean} over all $212$ instances for SigLIP-2, Chamfer Distance, and Uni3D~3D--3D. Because the evaluation set is fixed (failures contribute $0$; Chamfer uses a $1.5\!\times\!$max penalty), the $\Delta$ columns cleanly reflect overall benchmark improvement without the set-shift artifact that affects conditional-mean comparisons (where recovered instances enlarge the denominator).

Executability lifts substantially across all $22$ cells ($11$ models $\times$ $2$ tracks): $\sim\!90\%$ of baseline failures are Blender~5.0 API mismatches, which are localized and copy-pasteable as fixes once the traceback is visible. The penalized-mean SigLIP-2 delta is positive on every cell (image $+0.048$ to $+0.365$; text $+0.018$ to $+0.083$), confirming that multi-turn retry improves overall quality, not just executability. Chamfer and Uni3D show a clear capacity split: Claude and GPT families, which recover with high-quality meshes, show large Chamfer improvements ($-0.05$ to $-0.25$) and Uni3D gains ($+0.03$ to $+0.14$); Gemini Flash-class and Gemma models, whose recovered meshes tend to be simpler, show near-zero 3D-shape deltas despite large executability gains.

\textbf{Cost.}
Total dollar cost on baseline-failed instances is \textbf{\$55.54} (Gemma is free; GPT-5.4-mini and GPT-5.5 dominate at $\sim\!\$15$ each due to large failure pools at thinking=\texttt{medium}/\texttt{high}; Claude rows are cheapest at $\$1.1$--$\$2.3$). Wall-clock at four parallel workers ranges from $\sim\!10$~minutes on the largest Claude cell to $\sim\!2$~hours on Gemma~4~26B text.

% Multi-turn error-feedback retry — penalized-mean before/after.
\begin{table}[!htbp]
\centering
\caption{\textbf{Multi-turn error-feedback retry} ($T{=}3$ attempts). Left: executability lift (single-turn $\to$ multi-turn) with the recovery decomposition --- \emph{a$_0$ lucky} (fresh resample, no feedback) vs.\ \emph{mt fixed} (traceback consumed). Right: change in \emph{penalized mean} over all $212$ instances (failures contribute~$0$; Chamfer uses a $1.5\!\times\!$max penalty); because the evaluation set is fixed, $\Delta$ cleanly reflects overall benchmark improvement without set-shift artifacts. Per-row thinking level: \texttt{high} for Gemini/Gemma/GPT-5.5, \texttt{medium} for GPT-5.4-mini/GPT-5.4, \texttt{low} for Claude. Cost is the total dollar amount for baseline-failed instances only.}
\label{tab:multi-turn-debug}
\scriptsize
\renewcommand{\arraystretch}{0.88}
\setlength{\tabcolsep}{2.5pt}
\resizebox{\textwidth}{!}{%
\begin{tabular}{l l ccc rrr rrr r}
\toprule
 & & \multicolumn{3}{c}{Executability}
 & \multicolumn{3}{c}{Recoveries}
 & \multicolumn{3}{c}{$\Delta$\,penalized mean}
 & \\
\cmidrule(lr){3-5} \cmidrule(lr){6-8} \cmidrule(lr){9-11}
Model & Task
 & ST\,$\uparrow$
 & MT\,$\uparrow$
 & $\Delta$\,pp
 & a$_0$
 & mt
 & fail
 & SigLIP-2\,$\uparrow$
 & Chamfer\,$\downarrow$
 & Uni3D\,$\uparrow$
 & Cost\,\$ \\
\midrule
Gemini~3 Flash         & text  & 0.608 & 0.925 & $+$31.6 & 34 & 33 & 16 & $+$0.050 & $+$0.000 & $-$0.001 & 2.71 \\
Gemini~3 Flash         & image & 0.486 & 0.948 & $+$46.2 & 59 & 39 & 10 & $+$0.365 & $-$0.018 & $+$0.013 & 3.16 \\
Gemini~3.1 Flash Lite  & text  & 0.608 & 0.925 & $+$31.6 & 21 & 46 & 16 & $+$0.046 & $+$0.000 & $-$0.000 & 0.48 \\
Gemini~3.1 Flash Lite  & image & 0.552 & 0.934 & $+$38.2 & 36 & 45 & 14 & $+$0.282 & $+$0.000 & $+$0.001 & 0.67 \\
Gemini~3.1 Pro         & text  & 0.693 & 0.991 & $+$29.7 & --- & --- & 2  & $+$0.051 & $-$0.192 & $+$0.140 & --- \\
Gemini~3.1 Pro         & image & 0.703 & 0.995 & $+$29.2 & --- & --- & 1  & $+$0.239 & $-$0.194 & $+$0.120 & --- \\
Gemini~3.5 Flash       & text  & 0.410 & 0.910 & $+$50.0 & --- & --- & 19 & $+$0.083 & $+$0.035 & $+$0.218 & --- \\
Gemini~3.5 Flash       & image & 0.547 & 0.981 & $+$43.4 & --- & --- & 4  & $+$0.320 & $+$0.038 & $+$0.207 & --- \\
Gemma~4 26B            & text  & 0.467 & 0.906 & $+$43.9 & 48 & 45 & 19 & $+$0.072 & $-$0.004 & $+$0.003 & --- \\
Gemma~4 26B            & image & 0.604 & 0.948 & $+$34.4 & 37 & 36 & 11 & $+$0.271 & $+$0.000 & $-$0.001 & --- \\
Gemma~4 31B            & text  & 0.547 & 0.967 & $+$42.0 & 51 & 37 & 8  & $+$0.070 & $-$0.001 & $-$0.002 & --- \\
Gemma~4 31B            & image & 0.561 & 0.991 & $+$43.0 & 48 & 43 & 2  & \textbf{$+$0.339} & $+$0.003 & $+$0.000 & --- \\
Claude Sonnet~4.6      & text  & 0.802 & 0.986 & $+$18.4 & 27 & 12 & 3  & $+$0.027 & $-$0.152 & $+$0.080 & 2.26 \\
Claude Sonnet~4.6      & image & 0.882 & \textbf{1.000} & $+$11.8 & 17 & 8  & 0  & $+$0.110 & $-$0.109 & $+$0.036 & 1.52 \\
Claude Opus~4.7        & text  & 0.925 & \textbf{1.000} &  $+$7.5 & 11 & 5  & 0  & $+$0.018 & $-$0.116 & $+$0.070 & 1.28 \\
Claude Opus~4.7        & image & 0.948 & \textbf{1.000} &  $+$5.2 &  7 & 4  & 0  & $+$0.048 & $-$0.053 & $+$0.041 & 1.12 \\
GPT-5.4 mini           & text  & 0.670 & 0.991 & $+$32.1 & 51 & 17 & 2  & $+$0.054 & $-$0.221 & $+$0.140 & 11.25 \\
GPT-5.4 mini           & image & 0.792 & \textbf{1.000} & $+$20.8 & 38 & 6  & 0  & $+$0.166 & $-$0.253 & $+$0.109 & 3.91 \\
GPT-5.4                & text  & 0.863 & \textbf{1.000} & $+$13.7 & 21 & 8  & 0  & $+$0.024 & $-$0.120 & $+$0.068 & 5.84 \\
GPT-5.4                & image & 0.868 & \textbf{1.000} & $+$13.2 & 22 & 6  & 0  & $+$0.107 & $-$0.115 & $+$0.099 & 5.91 \\
GPT-5.5                & text  & 0.915 & \textbf{1.000} &  $+$8.5 & 16 & 2  & 0  & $+$0.029 & $-$0.064 & $+$0.026 & 11.31 \\
GPT-5.5                & image & 0.972 & \textbf{1.000} &  $+$2.8 &  5 & 1  & 0  & $+$0.051 & $-$0.194 & $+$0.145 & 4.12 \\
\bottomrule
\end{tabular}%
}
\end{table}

% -------------------------------------------------------------------------
\subsection{Visual Self-Critique Loop}
\label{app:visual-feedback}

\textbf{Setup.}
For each \emph{baseline-OK} instance, the model is shown its previous code, the four turntable renders it produced, and (image-to-3D only) the four reference images, then asked to either accept the result (\texttt{NEEDS\_FIX:NO}) or emit a corrected full Python script (\texttt{NEEDS\_FIX:YES} + \texttt{<assessment>} + \texttt{<code>}). Up to two iterations. The image-to-3D track uses an image-specific system prompt that explicitly frames the task as ``your renders vs.\ the reference images attached'' plus a \emph{revert-on-break} guard: if a fix's render fails, we restore the prior good state, so the loop is do-no-harm with respect to executability. Both system prompts and the user templates are in Appendix~\ref{app:prompt-visualfb}.

\textbf{Findings.}
Visual feedback is task-asymmetric (Table~\ref{tab:visual-feedback}). On text-to-3D it is uniformly positive across all four backbones ($+0.003$ to $+0.009$ mean $\Delta$\,SigLIP-2, win/lose ratios $1.24$--$2.63$, Gemma~4~26B strongest at $50/19$ wins/losses). On image-to-3D the same backbones flip uniformly negative ($-0.006$ to $-0.009$, win/lose ratios $0.58$--$0.78$). The per-model rank is similar across the two tasks, so the flip is task-fundamental, not model-fundamental: the image-task SigLIP-2 baseline is already at $0.78$--$0.81$ (close to the ceiling that voxel-style Infinigen renders attain), leaving little room to lift but plenty of room to introduce texture/lighting drift, and the metric weights global appearance over fine geometric correctness, so a geometrically improved fix that changes a leg's silhouette can score as a loss. We confirmed this by inspecting a sample of ``losses'' and finding several visually better fixes scoring lower; a stricter geometric metric (Chamfer, Uni3D~3D--3D) would likely re-flip the sign.

The DONE@1 acceptance rate on the image-to-3D rows is sharply capacity-ordered: Gemma~4~31B accepts $49/119$ baselines ($41\%$) versus $3$--$15$ ($3$--$13\%$) on the three smaller backbones, while reverts trend the opposite way ($11\%$ on $31$B vs.\ $26$--$28\%$ on smaller). The strongest backbone is the most conservative critic, and the smallest backbones are the most aggressive --- exactly the failure mode revert-on-break exists to absorb.

\begin{table}[htbp]
\centering
\caption{\textbf{Visual self-critique loop} (up to $2$ iterations on each baseline-OK instance). $n$ counts the baseline-OK instances on which the loop ran. \emph{$\Delta$\,SigLIP} is the per-instance post-loop minus baseline SigLIP-2 cosine, against the text description on the text-to-3D rows and against the reference views on the image-to-3D rows. Full setup in Appendix~\ref{app:visual-feedback}.}
\label{tab:visual-feedback}
\small
\setlength{\tabcolsep}{4pt}
\begin{tabular}{l l c rr r r}
\toprule
Model & Task & $n$
 & DONE@1
 & FIX/FIX
 & $\Delta$ siglip
 & win/lose \\
\midrule
\multicolumn{7}{l}{\textbf{Text-to-3D}} \\
\midrule
Gemini 3 Flash         & text  & 129 & 0   & 86 & $+$0.0030 & 46/31 \\
Gemini 3.1 Flash Lite  & text  & 129 & 2   & 85 & $+$0.0057 & 36/29 \\
Gemini 3.5 Flash       & text  &  75 & 8   & 39 & $+$0.0062 & 36/18 \\
Gemma 4 26B            & text  & 99  & 2   & 72 & \textbf{$+$0.0085} & \textbf{50/19} \\
Gemma 4 31B            & text  & 116 & 15  & 71 & $+$0.0055 & 47/34 \\
\midrule
\multicolumn{7}{l}{\textbf{Image-to-3D}} \\
\midrule
Gemini 3 Flash         & image & 103 & 3   & 86 & $-$0.0062 & 36/46 \\
Gemini 3.1 Flash Lite  & image & 117 & 15  & 78 & $-$0.0081 & 32/39 \\
Gemini 3.5 Flash       & image & 101 & 14  & 67 & \textbf{$+$0.0126} & \textbf{40/20} \\
Gemma 4 26B            & image & 128 & 6   & 93 & $-$0.0063 & 34/48 \\
Gemma 4 31B            & image & 119 & 49  & 39 & $-$0.0088 & 21/36 \\
\bottomrule
\end{tabular}
\end{table}

% -------------------------------------------------------------------------
\subsection{Text-to-Image-to-3D Pipeline}
\label{app:text-to-image-to-3d}

\textbf{Setup.}
We test whether inserting a strong text-to-image model as an intermediate ``visual grounding'' step helps text-to-3D. The intermediate is Nano Banana Pro (\texttt{gemini-3-pro-image-preview}), prompted with a clean studio three-quarter view template; we generate one photo per instance (\$$0.134$ each, \$$28.27$ for the full $212$). Five code-generators (Gemini~3~Flash, Flash~Lite, $3.1$~Pro, Gemma~4~26B, $31$B) are evaluated under three configurations: \textbf{A.\ direct text-to-3D} (original description $\to$ code-generator); \textbf{B.\ image-only} (the nbp photo $\to$ code-generator under the image-to-3D system prompt, $1$ view); \textbf{C.\ combined} (text + nbp photo together). All three are scored by SigLIP-2 text$\leftrightarrow$image cosine similarity against the \emph{original} description, so a higher score indicates greater alignment with the user's stated intent, regardless of which intermediate the pipeline used.

\begin{table}[htbp]
\centering
\caption{\textbf{Text-to-image-to-3D pipeline.} \textbf{A.~direct}: text $\to$ code-generator (headline text-to-3D); \textbf{B.~image-only}: text $\to$ Nano Banana Pro $\to$ single reference image $\to$ image-task code-generator; \textbf{C.~combined}: B with the original text prepended to the image-task user message. All three are scored by SigLIP-2 cosine against the \emph{original text description}; SigLIP columns are penalized means (failed renders contribute $0$). $\Delta$ columns are signed differences vs.\ A. Full setup in Appendix~\ref{app:text-to-image-to-3d}.}
\label{tab:text-to-image-pipeline}
\small
\setlength{\tabcolsep}{4pt}
\begin{tabular}{l ccc ccc rr}
\toprule
Model
 & A.exec
 & B.exec
 & C.exec
 & A.SigLIP
 & B.SigLIP
 & C.SigLIP
 & $\Delta_{B-A}$
 & $\Delta_{C-A}$ \\
\midrule
Gemini 3 Flash         & 0.81 & 0.54 & 0.56 & 0.142 & 0.083 & 0.085 & $-$0.059 & $-$0.053 \\
Gemini 3.1 Flash Lite  & 0.84 & 0.48 & 0.49 & 0.130 & 0.063 & 0.065 & $-$0.067 & $-$0.059 \\
Gemini 3.1 Pro         & 0.79 & 0.75 & \textbf{0.85} & 0.140 & 0.122 & \textbf{0.144} & $-$0.018 & \textbf{$+$0.011} \\
Gemini 3.5 Flash       & 0.41 & 0.52 & 0.54 & 0.082 & 0.055 & 0.094 & $-$0.027 & $+$0.012 \\
Gemma 4 26B            & 0.61 & 0.72 & 0.78 & 0.097 & 0.102 & \textbf{0.113} & $+$0.006 & \textbf{$+$0.014} \\
Gemma 4 31B            & 0.69 & 0.75 & 0.81 & 0.120 & 0.119 & \textbf{0.123} & $-$0.001 & \textbf{$+$0.005} \\
\bottomrule
\end{tabular}
\end{table}

\textbf{Findings.} Image-only (B) regresses against direct text-to-3D on $4$ of $5$ backbones, and the loss is biggest on the smaller Flash class ($-0.059$, $-0.067$); even $3.1$~Pro drops $-0.018$. Two effects compound: a single $1408\times 768$ photo discards specific text constraints (``boxy frame with two raised armrests'') into diffuse visual hints, and the photorealistic style biases the code-generator toward more elaborate geometry than the small models can write without crashing (executability falls $27$~pp on Gemini~3~Flash from \emph{A} to \emph{B}). The combined mode (C) re-anchors generation on the original text and recovers most of the loss; on the three high-capacity / high-thinking backbones (Pro, Gemma~$26$B, Gemma~$31$B) it \emph{flips positive} over direct text-to-3D ($+0.011$, $+0.014$, $+0.005$), while the two Flash backbones remain net negative because their thinking budget cannot reconcile photo and text without overflowing. Practically, this extension pays off when the backbone is strong enough to handle multimodal inputs and the text description carries details not recoverable from a single rendered view; the marginal cost is a $\sim\!\$0.13$ per-instance nbp call.

% =========================================================================
\section{3DCodeArena: human-preference voting interface}
\label{app:arena}

To complement the automated metric suite of Section~\ref{app:metrics}, we operate a public LMSYS-style pairwise-vote arena (URL withheld for anonymous review; the live deployment, schema, and front-end source are included in the supplemental material).
The arena is a thin Next.js front-end backed by a Supabase Postgres database; each vote is keyed by a stable browser fingerprint to suppress trivial duplicates, and the underlying \texttt{votes} table is the same one we snapshot for the auto-judge study of Appendix~\ref{app:llm-judge}.
At the time of writing, the arena hosts $12$ frontier VLMs across both modality tracks and has collected roughly $2{,}500$ human votes; per-model Elo ratings (Bradley--Terry MLE, recentered to mean $1000$ and converted to Elo points at $400/\ln 10$ per logit, with $1000$-resample bootstrap $95\%$ CIs) are recomputed nightly and surfaced on a public leaderboard.

\textbf{Blind pairwise comparison.}
On every visit the server samples a prompt from the canonical $212$-instance \ourdata{} set, picks two distinct model variants that have produced an executable mesh for that prompt, and serves their GLB exports side by side as \emph{Model A} and \emph{Model B} --- the model identities are never revealed until after the vote.
Both viewers re-skin every loaded mesh with a single shared neutral-gray \texttt{MeshStandardMaterial} (\texttt{color~=~0xb8b8bc}, \texttt{roughness~=~0.7}, \texttt{metalness~=~0}, double-sided to mask flipped-normal artifacts).
Stripping textures and colors forces voters to judge geometry alone, in line with our metric protocol of scoring untextured renders.
Cameras auto-fit the mesh's bounding box on load, and the \texttt{OrbitControls} rig lets voters drag-rotate and scroll-zoom each side independently; presentation order is randomized per vote with a deterministic swap bit so any positional bias averages over A/B identity.
After inspecting the pair, the voter picks one of four buttons --- \emph{A is better}, \emph{B is better}, \emph{Tie}, or \emph{Both bad} --- and the server records the verdict with the un-swapped model identities, model display names, the prompt slug, and the voter fingerprint.

\begin{figure}[!ht]
\centering
\includegraphics[width=\linewidth]{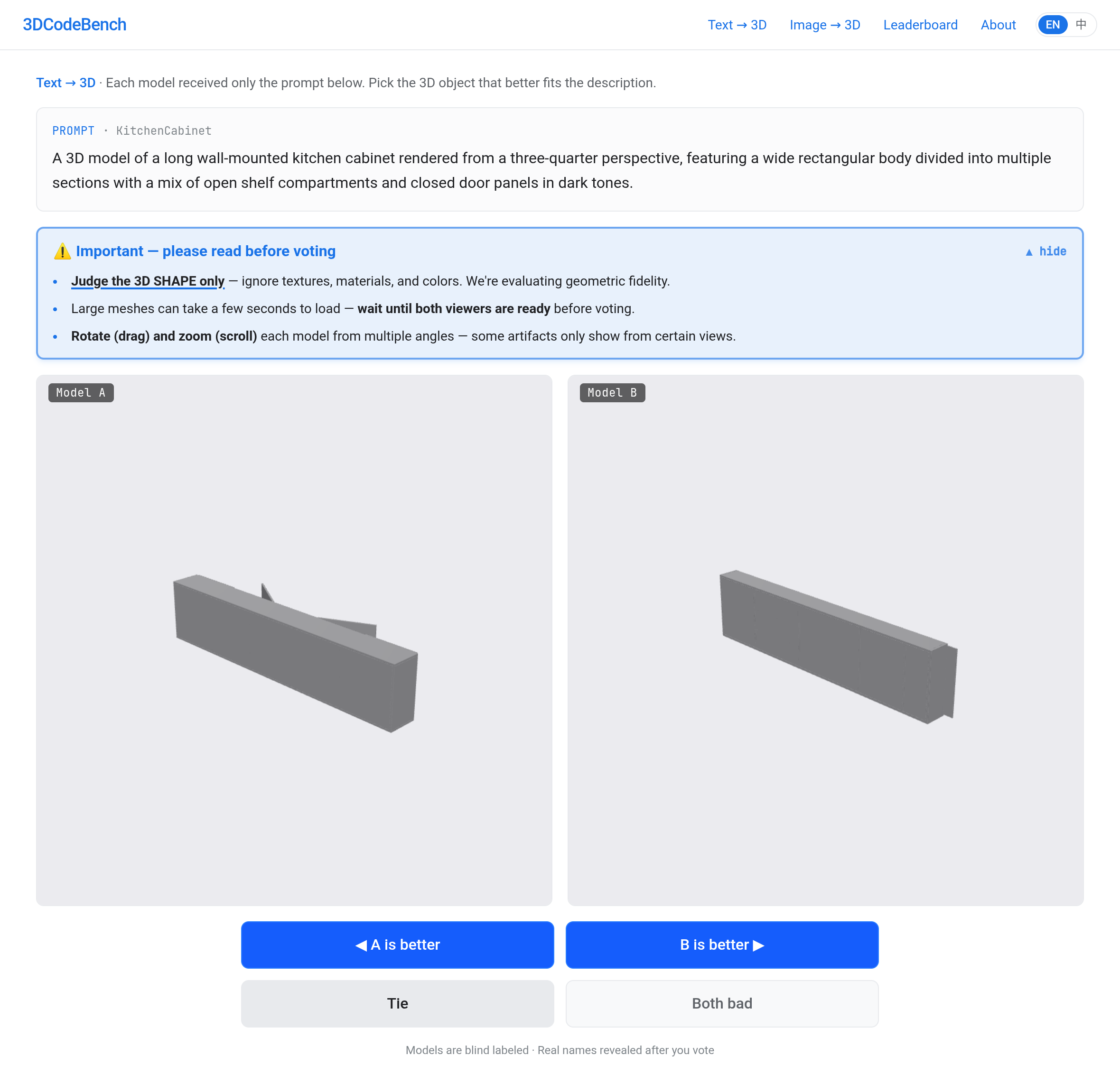}
\caption{\textbf{3DCodeArena --- Text-to-3D interface.} Voters see the natural-language prompt at the top and two anonymized gray-shaded GLB renders side by side; both viewers are independently orbit-rotatable and zoomable, and the vote is cast with one of four buttons (\emph{A is better} / \emph{B is better} / \emph{Tie} / \emph{Both bad}). The pairing shown here is real (a perching-bird prompt; both meshes are the live exports our voters saw); model identities are revealed only after the vote.}
\label{fig:arena-text}
\end{figure}

\begin{figure}[!ht]
\centering
\includegraphics[width=\linewidth]{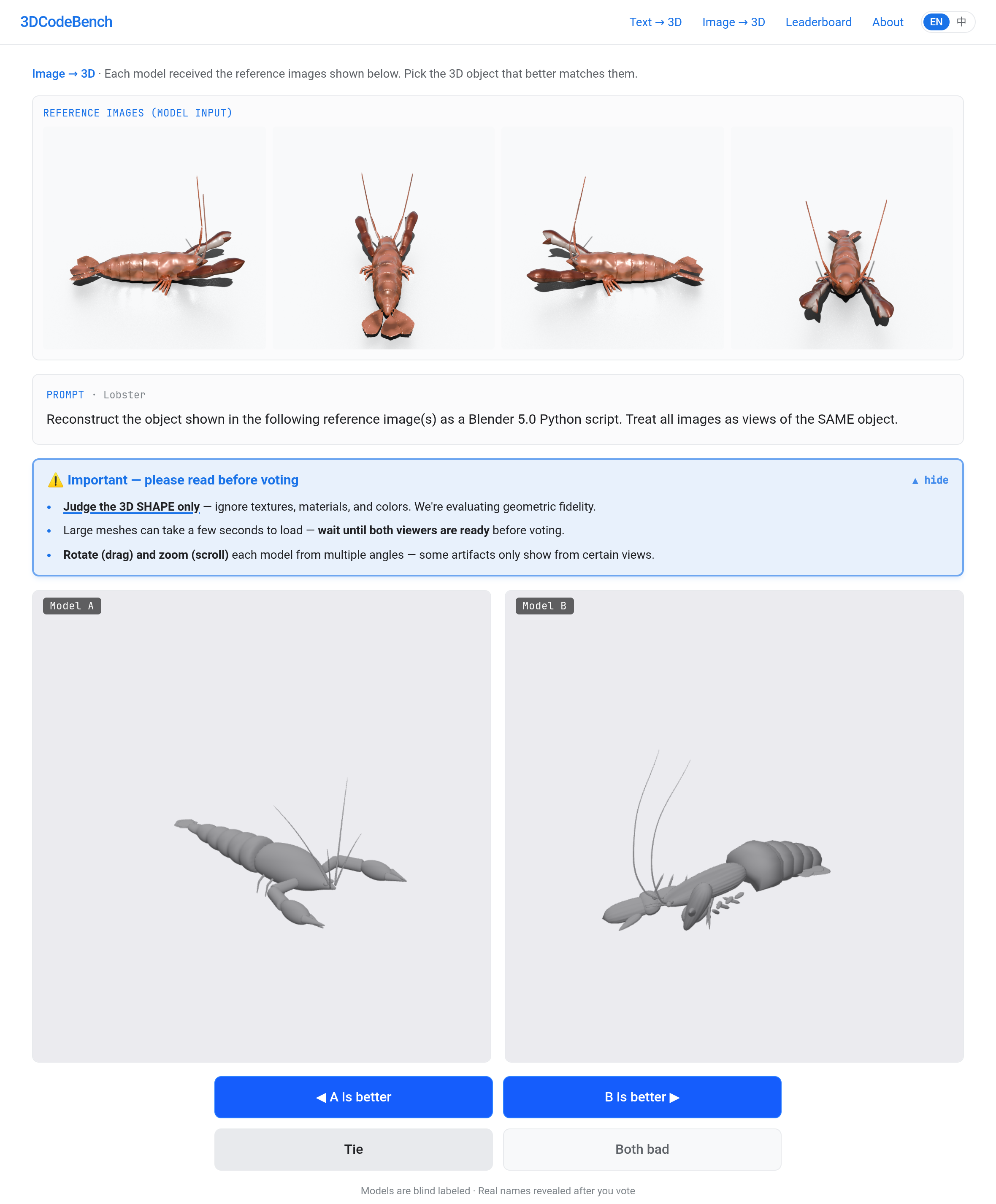}
\caption{\textbf{3DCodeArena --- Image-to-3D interface.} The same blind pairwise layout as Figure~\ref{fig:arena-text}, with the additional reference-image strip at the top: the four canonical views of the target object that the policy was conditioned on are rendered above the prompt so voters can score multi-view faithfulness directly against the input. The pairing shown here is a tube-coral prompt; voters again judge only the geometry.}
\label{fig:arena-image}
\end{figure}

\textbf{Per-track interfaces.}
Figure~\ref{fig:arena-text} shows the \emph{Text-to-3D} track: voters see only the natural-language prompt above the two viewers, mirroring the input the policy received.
Figure~\ref{fig:arena-image} shows the \emph{Image-to-3D} track: the same blind pairwise layout, with a four-image reference strip rendered above the prompt so voters can compare the two meshes directly against the target views the model conditioned on.
Per-modality Elo is kept on a separate scale because the two tracks exercise different skills (long-form geometric reasoning from text vs.\ multi-view-grounded reconstruction) and conflating them would mix qualitatively different signals.
A persistent collapsible ``Important --- please read before voting'' tip box reminds voters of three rules that we found materially affect vote quality in pilot runs: (i) judge \emph{shape} only, ignoring textures, materials, and colors; (ii) wait until \emph{both} viewers finish loading before voting, since large meshes can take several seconds; (iii) \emph{rotate and zoom} each model from multiple angles, since some artifacts only appear from particular viewpoints.

\textbf{Vote-pool reuse.}
The same \texttt{votes} table that drives the live leaderboard is the human ground truth against which the LLM/VLM-as-a-judge study of Appendix~\ref{app:llm-judge} is scored; a snapshot of $n{=}2{,}508$ vote rows (after dropping a small fraction with missing local artifacts) forms the four-verdict alphabet on which we compute exact-match agreement and decisive-row accuracy for the four Google judges in image and code modes.
The public arena URL, the deployed front-end, and the Supabase schema are all included in the released \texttt{eval/arena/} directory.

\begin{figure}[!htbp]
\centering
\begin{subcaptionblock}{0.82\linewidth}
\centering
\includegraphics[width=\linewidth]{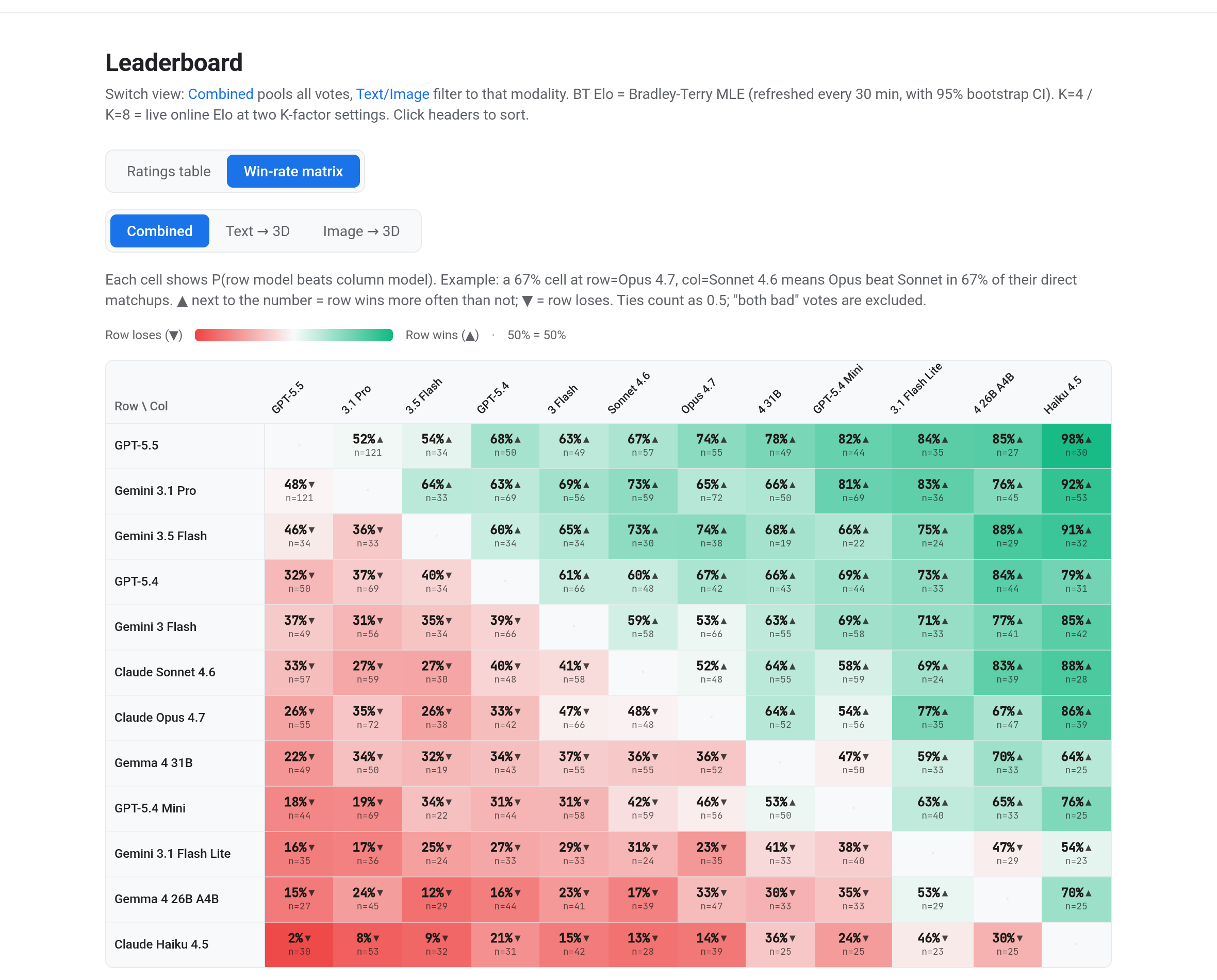}
\caption{Combined (both tracks pooled).}
\label{fig:arena-winrate-combined}
\end{subcaptionblock}\\[4pt]
\begin{subcaptionblock}{0.42\linewidth}
\centering
\includegraphics[width=\linewidth]{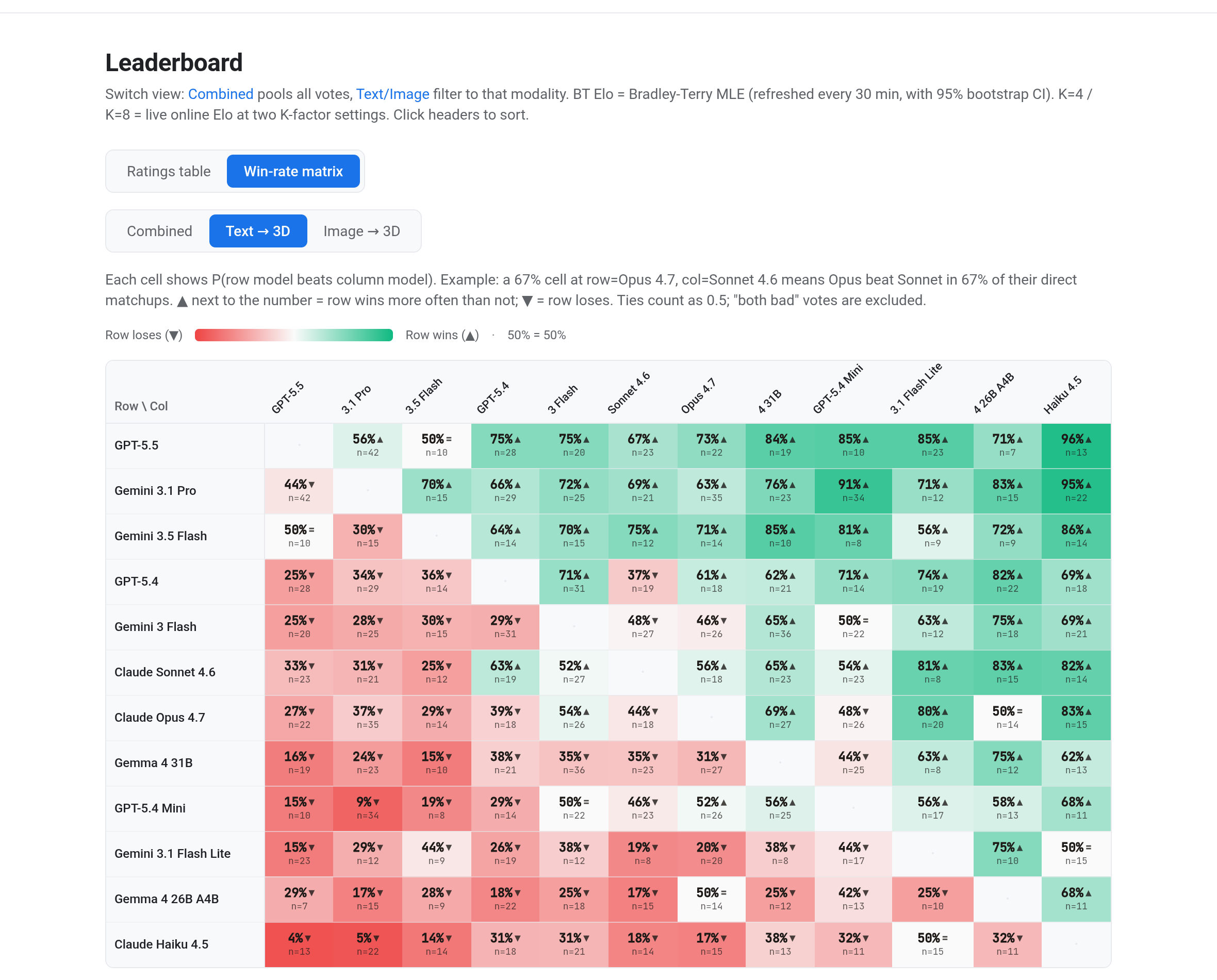}
\caption{Text-to-3D.}
\label{fig:arena-winrate-text}
\end{subcaptionblock}
\hfill
\begin{subcaptionblock}{0.42\linewidth}
\centering
\includegraphics[width=\linewidth]{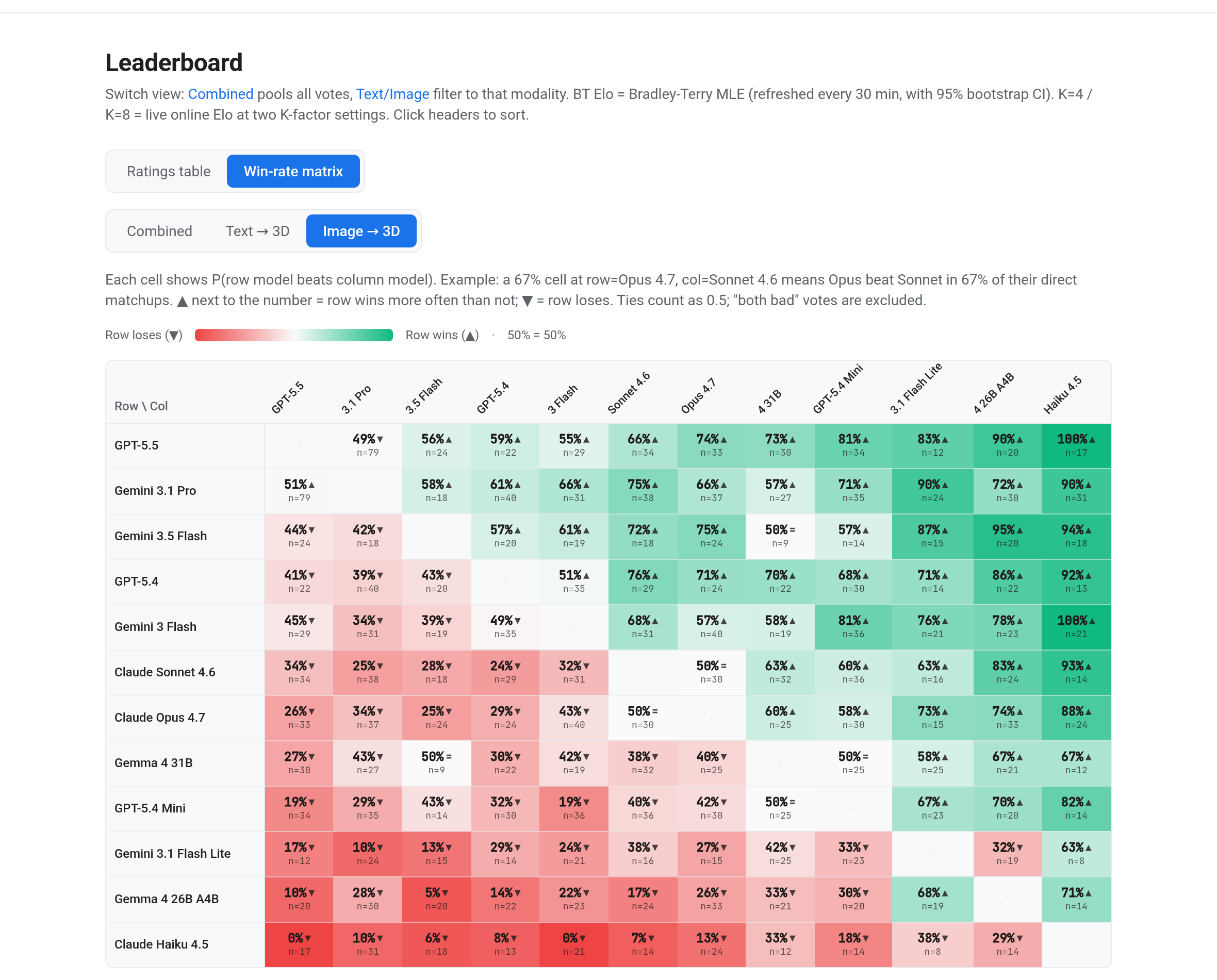}
\caption{Image-to-3D.}
\label{fig:arena-winrate-image}
\end{subcaptionblock}
\caption{\textbf{3DCodeArena --- live win-rate matrix.} Each cell is $P(\text{row beats column})$ on directly observed head-to-head votes; $n$ below the percentage is the per-pair sample size; ties and ``both bad'' contribute $0.5$/$0.5$. Models are sorted in row/column order by overall Elo, so the green upper-triangle / red lower-triangle pattern is the visual signature of the leaderboard ranking, with cells closer to white indicating pairings the human pool has not yet decisively separated.}
\label{fig:arena-winrate}
\end{figure}

\textbf{Live leaderboard and win-rate matrix.}
Beyond the per-vote interface, the public site exposes a leaderboard view that aggregates the same \texttt{votes} table into both Bradley--Terry Elo (recomputed every $30$~min, with $95\%$ bootstrap CI; $K{=}4$ and $K{=}8$ live online variants) and a per-pair win-rate matrix. The matrix view (Figure~\ref{fig:arena-winrate}) reports $P(\text{row beats column})$ on every directly observed pairing, with cell color saturation tracking distance from the $50\%$ no-preference line, the per-cell sample size $n$ printed below the percentage, and ties / both-bad counted as $0.5$/$0.5$. The three sub-tabs (Combined / Text$\to$3D / Image$\to$3D) let readers inspect the same $12$-model field under each modality slice and cross-check whether a given Elo gap is supported by enough head-to-head votes to be meaningful.

\section{LLM/VLM-as-a-judge against the human arena}
\label{app:llm-judge}

We ran a controlled study to ask whether a frontier LLM/VLM, given the same pairwise comparison the human voters saw, can recover human verdicts --- both as a render-based proxy for the Bradley--Terry signal and as a cheaper code-only auto-judge that skips Blender entirely.

\subsection{Data and Protocol}
\label{app:judge-protocol}

\textbf{Vote pool.}
We snapshotted the live arena \texttt{votes} table ($2{,}560$ votes across the $212$-instance \ourdata{} set, both tracks). For each vote, we resolved (a) four canonical renders per side at frames $\{5, 15, 25, 35\}$ (matching the SigLIP-2 metric's camera angles) and (b) the original Blender Python source. After dropping $52$ votes ($2\%$) with missing local artifacts, the judge set has $n{=}2{,}508$ pairs distributed as $1{,}012$~\texttt{a}\,/\,$1{,}105$~\texttt{b}\,/\,$192$~\texttt{tie}\,/\,$199$~\texttt{both\_bad} ($1{,}455$ image-track, $1{,}053$ text-track).

\textbf{Judges and modes.}
Four open-API Google models judge every pair --- Gemini~3 Flash, Gemini~3.1 Flash Lite, Gemini~3.1 Pro, Gemma~4 31B --- each in two modes: \emph{image} ($4$ renders/side $+$ the $1{-}4$ reference images on image-track prompts) and \emph{code} (full \texttt{.py} source/side, $60$\,K-character cap, $<$$1\%$ of pairs hit it). Sampling is greedy ($T{=}0$) at thinking=\texttt{low} (Gemma has no tunable thinking). Model identities are anonymized as System~A / System~B, and the response is a strict single-line JSON with \texttt{winner} $\in \{a, b, \text{tie}, \text{both\_bad}\}$ and a one-to-three-sentence \texttt{reasoning} field.

\textbf{Position-bias control.}
For each pair, a deterministic per-vote swap bit (\texttt{Random(vote\_id)\allowbreak{}.\allowbreak{}random()}\,$<$\,$0.5$) decides whether to exchange the two sides' content before formatting; the verdict is unswept after the call. With approximately balanced swap, any positional preference averages out over A/B identity and registers only as residual symmetric noise.

\textbf{Cost.}
The full sweep is $4 \times 2 \times 2{,}508 \approx 20$\,K API calls in $\sim\!4$ wall-clock hours; $13$ pairs across all eight cells failed JSON parsing and were dropped.

\subsection{Judge Prompt}
\label{app:judge-prompt}

The image-mode system prompt (\texttt{eval/judge/prompts/image\_judge.txt}, abridged):

\begin{spromptbox}[Judge system prompt --- image mode]{spromptGray}{spromptGrayBg}
\small
You are an impartial judge in a 3D-modeling arena. Two anonymous AI systems were each given the SAME prompt and asked to produce a 3D object in Blender using Python. We rendered the resulting 3D object from four canonical viewpoints. Your job is to decide which side better matches the prompt, judging only from the renders.\dots

\medskip
\noindent\textbf{\# Verdict options}\\
\texttt{a} / \texttt{b} / \texttt{tie} / \texttt{both\_bad}.

\medskip
\noindent\textbf{\# Criteria (weighted in order)}
\begin{enumerate}[leftmargin=1.5em,itemsep=1pt,topsep=2pt]
  \item Object identity.
  \item Structural correctness.
  \item Geometric detail.
  \item For image-to-3D: faithfulness to the reference images.
\end{enumerate}
Color/material is NOT a criterion.

\medskip
\noindent\textbf{\# Output format}\\
Output STRICTLY a single JSON object \texttt{\{"winner": ..., "reasoning": ...\}}.
\end{spromptbox}
The code-mode prompt is structurally parallel but instructs the judge to assess each script's likelihood of producing the correct object \emph{without running it} (object intent, structural decomposition, geometric detail, executability under Blender~5.0). Both prompts share the strict JSON contract; full templates are in \texttt{eval/judge/prompts/}.

\subsection{Headline Results: The Four-Verdict View}
\label{app:judge-results-full}

\begin{table}[htbp]
\centering
\caption{LLM/VLM-as-a-judge agreement with the human arena, four-verdict
formulation. \emph{Agree} is the exact-match rate over all $n$ decisive
+ non-decisive votes; \emph{Decisive} restricts to the rows where the
human picked \texttt{a} or \texttt{b} (\textbf{judge} verdicts of
\texttt{tie}/\texttt{both\_bad} count as wrong). Bold marks the best
judge per mode.}
\label{tab:judge-full}
\small
\setlength{\tabcolsep}{4.5pt}
\begin{tabular}{l l r r r r r r r r r}
\toprule
Judge & Mode & $n$ & Agree & Decisive & img-acc & txt-acc
       & $a$ & $b$ & tie & b-bad \\
\midrule
\multicolumn{11}{l}{\emph{Image (renders) judging}} \\
\midrule
Gemini 3.1 Pro              & image & 2508 & \textbf{64.7\%} & 72.6\% & 65.4\% & 63.7\% & 1129 & 1162 &  64 & 153 \\
Gemini 3 Flash              & image & 2508 & 64.2\%          & 74.4\% & 64.3\% & 64.1\% & 1183 & 1248 &  34 &  43 \\
Gemini 3.1 Flash Lite       & image & 2508 & 63.0\%          & 73.4\% & 62.9\% & 63.1\% & 1173 & 1268 &  44 &  23 \\
Gemma 4 31B                 & image & 2508 & 62.5\%          & 72.6\% & 63.0\% & 61.9\% & 1137 & 1293 &  34 &  44 \\
\midrule
\multicolumn{11}{l}{\emph{Code (raw \texttt{.py}) judging}} \\
\midrule
Gemini 3 Flash              & code  & 2506 & \textbf{56.9\%} & 67.4\% & 56.2\% & 57.9\% & 1111 & 1395 &   0 &   0 \\
Gemma 4 31B                 & code  & 2505 & 55.4\%          & 65.6\% & 55.1\% & 55.8\% & 1078 & 1425 &   0 &   2 \\
Gemini 3.1 Flash Lite       & code  & 2502 & 52.6\%          & 62.3\% & 52.9\% & 52.3\% & 1101 & 1387 &   6 &   8 \\
Gemini 3.1 Pro              & code  & 2506 & 51.7\%          & 59.6\% & 50.1\% & 53.9\% &  955 & 1234 & 317 &   0 \\
\bottomrule
\end{tabular}
\end{table}

\textbf{Image judging is usable; code judging is borderline.}
Image judges land at $62.5\%{-}64.7\%$ overall agreement (decisive subset $\sim\!73\%$), well above the $25\%$ four-class chance line and the $44.1\%$ majority baseline. Code judges drop $\sim\!7{-}13$\,pp to $51.7\%{-}56.9\%$ ($59.6\%{-}67.4\%$ decisive) --- usable for trend-level comparisons but not interchangeable with renders.

\textbf{Pro alone uses \texttt{tie} and \texttt{both\_bad}.}
Flash, Flash Lite, and Gemma collapse to a binary $a/b$ vote: non-decisive verdicts make up $0\%{-}3.1\%$ of calls vs.\ the human rate of $15.6\%$. Pro abstains on $8.7\%$ of image and $12.6\%$ of code calls. The four-verdict exact-match metric punishes this calibration (every Pro \texttt{tie} on a decisive human row counts as wrong), which is most of why Pro looks worst on Table~\ref{tab:judge-full} despite leading the decisive subset.

\subsection{A/B-Only View: Accuracy and Correlation on Decisive--Decisive Rows}
\label{app:judge-results-ab}

Dropping \texttt{tie} and \texttt{both\_bad} on \emph{both} sides isolates the question \emph{when human and judge both committed, how often did they agree?} on $1{,}864{-}2{,}116$ doubly-decisive rows per cell. We additionally report Cohen's $\kappa$ (chance-corrected on the binary alphabet) and Pearson $\phi$ on $a{=}0, b{=}1$ for cross-cell strength comparison.

\begin{table}[htbp]
\centering
\caption{A/B-only stats: drop \texttt{tie} and \texttt{both\_bad} on
\emph{both} the human and the judge side, then compute accuracy and
two correlation coefficients on the remaining decisive--decisive rows.
$\kappa$ is Cohen's kappa; $\phi$ is the Pearson coefficient on
$a{=}0,b{=}1$ (equivalent to phi on the $2{\times}2$ table).
\emph{Coverage} is the fraction of the approximately $2{,}117$ human-decisive votes
the judge also called decisively.}
\label{tab:judge-ab}
\small
\setlength{\tabcolsep}{4.5pt}
\resizebox{\textwidth}{!}{%
\begin{tabular}{l l r r r r r r r r r r r}
\toprule
Judge & Mode & $n_{ab}$ & Acc & $\kappa$ & $\phi$ & Cov
       & img-acc & txt-acc
       & \multicolumn{4}{c}{Confusion (h$\rightarrow$j)} \\
\cmidrule(lr){10-13}
& & & & & & & & & $aa$ & $ab$ & $ba$ & $bb$ \\
\midrule
\multicolumn{13}{l}{\emph{Image (renders) judging}} \\
\midrule
Gemini 3.1 Pro              & image & 1994 & \textbf{77.1\%} & \textbf{+0.542} & +0.543 & 94.2\% & 78.0\% & 76.0\% & 741 & 212 & 244 & 797 \\
Gemini 3 Flash              & image & 2082 & 75.6\%          & +0.513          & +0.513 & 98.3\% & 76.7\% & 74.3\% & 754 & 239 & 268 & 821 \\
Gemini 3.1 Flash Lite       & image & 2079 & 74.7\%          & +0.494          & +0.494 & 98.2\% & 75.5\% & 73.8\% & 739 & 253 & 272 & 815 \\
Gemma 4 31B                 & image & 2075 & 74.0\%          & +0.479          & +0.479 & 98.0\% & 75.3\% & 72.3\% & 718 & 272 & 267 & 818 \\
\midrule
\multicolumn{13}{l}{\emph{Code (raw \texttt{.py}) judging}} \\
\midrule
Gemini 3.1 Pro              & code  & 1864 & \textbf{67.7\%} & \textbf{+0.348} & +0.349 &  88.1\% & 68.8\% & 66.2\% & 543 & 330 & 273 & 718 \\
Gemini 3 Flash              & code  & 2116 & 67.4\%          & +0.345          & +0.345 & 100.0\% & 67.0\% & 67.9\% & 629 & 383 & 307 & 797 \\
Gemma 4 31B                 & code  & 2113 & 65.6\%          & +0.307          & +0.309 & 100.0\% & 65.7\% & 65.5\% & 593 & 417 & 310 & 793 \\
Gemini 3.1 Flash Lite       & code  & 2102 & 62.6\%          & +0.249          & +0.249 &  99.5\% & 63.6\% & 61.3\% & 579 & 426 & 360 & 737 \\
\bottomrule
\end{tabular}%
}
\end{table}

\textbf{Pro leads on image once the abstention penalty is stripped.}
The image ordering flips: Pro $77.1\%$ accuracy / $\kappa{=}{+}0.542$ (substantial on Landis--Koch) vs.\ Flash $75.6\%/{+}0.513$, Flash~Lite $74.7\%/{+}0.494$, Gemma $74.0\%/{+}0.479$. Pro's coverage of human-decisive rows is $94.2\%$ vs.\ $98\%{-}98.3\%$ for the others --- a $\sim\!4$\,pp throughput cost for the $\sim\!1.5$\,pp accuracy gain.

\textbf{Code: Pro and Flash tied, $\kappa$ in fair-to-moderate.}
A/B accuracy lands at $62.6\%{-}67.7\%$ with $\kappa{=}{+}0.249$ to ${+}0.348$. Pro and Flash are within $0.3$\,pp ($67.7\%$ vs.\ $67.4\%$); Flash~Lite trails meaningfully ($\kappa{=}{+}0.249$), with reasoning traces that weight surface-code style over geometric intent.

\textbf{Image-track prompts are easier than text-track for every image judge.}
The image-track gives judges four reference views as an additional anchor; image-mode accuracy is $2{-}3$\, pp higher on image-track than text-track rows (e.g.\ Pro $78.0\%$ vs.\ $76.0\%$). The gap vanishes in code mode, where references are unavailable, consistent with the gain arising from the visual anchor rather than from prompt-track difficulty.

% \clearpage
\section{Sampling temperature ablation}
\label{app:temperature-ablation}

\begin{figure}[!ht]
\centering
\includegraphics[width=\linewidth]{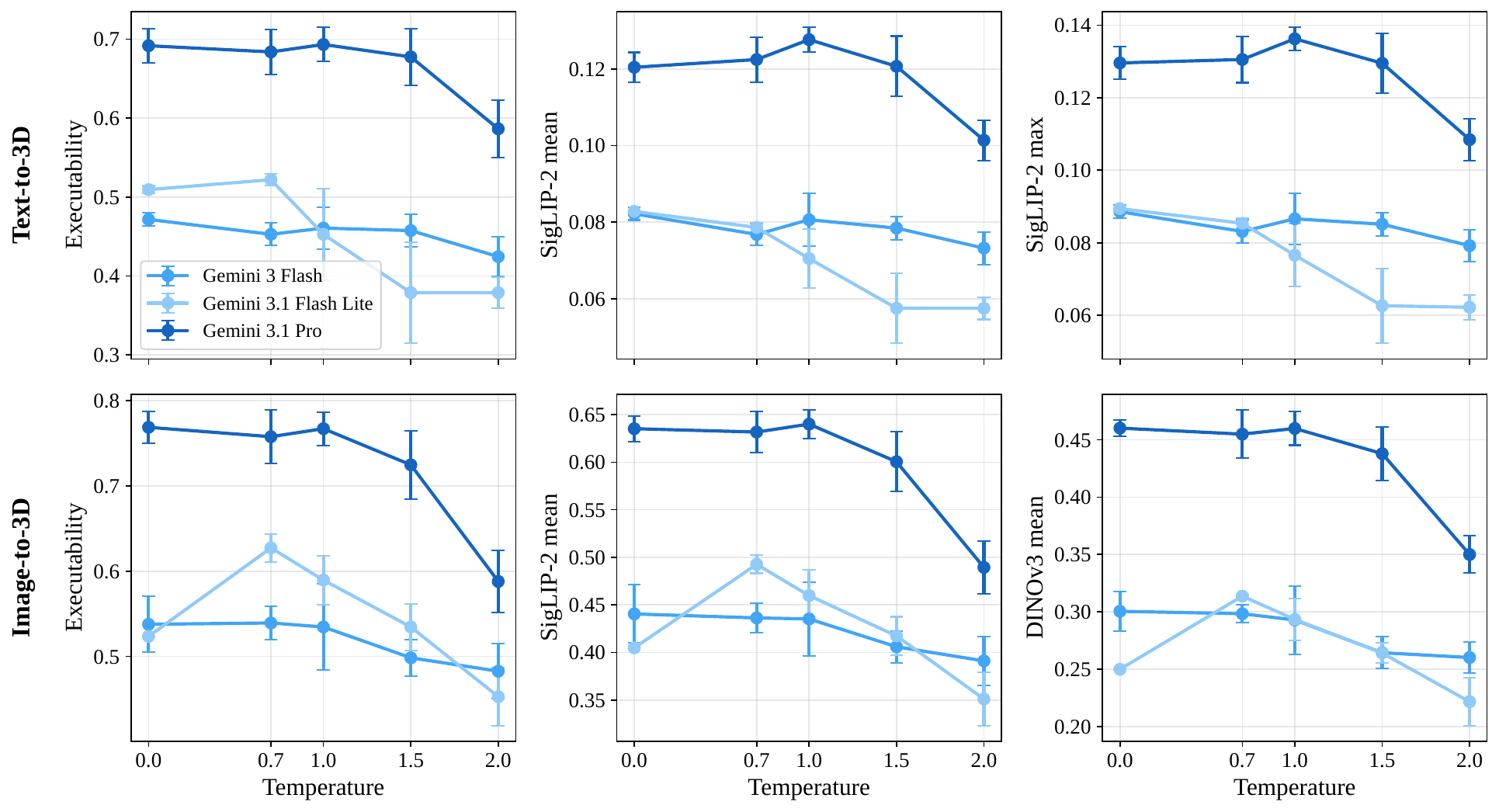}
\caption{\textbf{Quality vs.\ sampling temperature}, $3$ seeds per cell at \texttt{thinking\_level=high}, $1\sigma$ error bars across seeds. \emph{Top:} text-to-3D (Exec, SigLIP-2 mean, SigLIP-2 max). \emph{Bottom:} image-to-3D (Exec, SigLIP-2 mean, DINOv3 mean). Pro is largely flat for $T \in [0, 1.5]$ and only drops at $T{=}2.0$; Flash is similarly robust; Flash~Lite peaks sharply at $T{=}0.7$ and degrades above $T{=}1.0$. $T{=}2.0$ is the worst configuration in every cell.}
\label{fig:temp}
\end{figure}

The headline numbers in Table~\ref{tab:t1} use \texttt{temperature}\,$=0.7$. Figure~\ref{fig:temp} sweeps \texttt{temperature} $\in \{0, 0.7, 1.0, 1.5, 2.0\}$ on three Gemini~3 models (Flash, Flash~Lite, Pro) at \texttt{thinking\_level}\,$=$\,\texttt{high}, with $3$ seeds per cell at fixed values $\{0,1,2\}$ and all other settings at the paper's defaults; the five points cover greedy decoding, our baseline, Gemini's documented default, and two above-default settings up to the API maximum.

\textbf{$T{=} 2.0$ is uniformly worst; $\{0, 0.7, 1.0\}$ is indistinguishable on the strong models.}
On Flash and Pro, the three low-$T$ cells overlap within $1\sigma$ on every metric (e.g.\ Pro text-to-3D Exec $0.692/0.684/0.693$). Flash~Lite peaks at $T{=}0.7$ (image $0.627\!\pm\!0.016$) and collapses with a $\sim\!17$\,pp spread between $T{=}0.7$ and $T{=}2.0$. Pro is the most temperature-robust, losing only $\sim\!10/18$\,pp on text/image at $T{=}2.0$.

\textbf{Greedy decoding is expensive on Flash-class, mild on Pro.}
$T{=}0$ outputs are nearly bit-identical across seeds but \emph{thinking} is not, and inflates sharply on Flash-class: Flash~Lite text-to-3D burns $\sim\!63$\,K thinking tokens/instance (\$20.34 per $212$-instance pass) at $T{=}0$ vs.\ $\sim\!4$\,K (\$2.47) at $T{=}0.7$, matching Google's own warning against $T{<}1$. Pro is graceful: $T{=}0$ thinking is below $T{=}1.0$ thinking on both tracks. \textbf{Recommendation:} $T{=}0.7$ is on the Pareto frontier; we recommend against $T \geq 1.5$.

\end{document}